\documentclass{article} 
\usepackage{iclr2026_conference,times}

\usepackage{amsmath,amsfonts,bm}

\def\eqref#1{equation~\ref{#1}}

\def\1{\bm{1}}

\DeclareMathAlphabet{\mathsfit}{\encodingdefault}{\sfdefault}{m}{sl}
\SetMathAlphabet{\mathsfit}{bold}{\encodingdefault}{\sfdefault}{bx}{n}

\usepackage{hyperref}
\usepackage{url}
\usepackage{wrapfig}
\usepackage{pifont}
\usepackage{color}
\usepackage{booktabs} 
\usepackage{multirow} 
\usepackage{rotating} 
\usepackage{array} 
\usepackage{graphicx} 
\usepackage{caption} 
\usepackage{threeparttable} 
\usepackage{adjustbox} 
\usepackage{xcolor} 
\usepackage{siunitx}
\usepackage{makecell}
\usepackage[table]{xcolor} 
\usepackage{ragged2e} 
\usepackage{xspace}   
\usepackage{multibib}
\usepackage{enumitem}
\usepackage{amsmath}  
\newcommand{\cmarkb}{\textbf{\ding{51}}}
\newcommand{\xmarkb}{\textbf{\ding{55}}}
\usepackage{float}
\usepackage[ruled,vlined]{algorithm2e}
\usepackage{listings}
\usepackage{bbm}
\usepackage{etoc}

\title{UniPhysGen: Unified Physical Grounding for Simulation-Ready 3D Assets}

\author{
\parbox[t]{0.95\textwidth}{
Xian Li\textsuperscript{1,2}, 
Rong Wei\textsuperscript{2},
Lujie Yang\textsuperscript{2,3},
Haolin Huang\textsuperscript{1},
Junyuan Fang\textsuperscript{2},
Siliang Tang\textsuperscript{1},
Jun Xiao\textsuperscript{1},
Rui Tang\textsuperscript{2}\thanks{Project leader},
Juncheng Li\textsuperscript{1}\thanks{Corresponding author: junchengli@zju.edu.cn}
}\\
\parbox[t]{0.9\textwidth}{
\hspace*{0.3em}\textsuperscript{1} Zhejiang University, \textsuperscript{2} Manycore Tech Inc.\\
\hspace*{0.3em}\textsuperscript{3} University of Electronic Science and Technology of China}
}

\iclrfinalcopy 
\begin{document}

\maketitle

\begin{abstract}
Physically grounded 3D assets are increasingly important for embodied AI and robotic simulation. However, most existing 3D assets lack unified physical semantics, including articulation semantics and intrinsic physical properties, required for realistic interaction. Current approaches either treat these semantics independently or rely on canonicalized object structures, limiting robustness across heterogeneous 3D assets.
We present \textbf{UniPhys}, a scalable framework for automatically transforming raw 3D assets into simulation-ready assets with unified physical semantics. Based on UniPhys, we construct \textbf{UniPhys-40K}, a large-scale physically grounded dataset, together with \textbf{UniPhys-Bench}, a carefully verified benchmark for unified physical grounding evaluation.
We further introduce \textbf{UniPhysGen}, a unified physical grounding model that jointly reasons over articulation semantics and intrinsic physical properties. UniPhysGen incorporates geometry-robust articulation grounding to mitigate geometric shortcut bias under heterogeneous part decompositions. Extensive experiments demonstrate state-of-the-art performance across articulation grounding and intrinsic physical property estimation tasks, while the resulting assets can be directly deployed in robotic simulation environments for realistic physical interaction. Our code and dataset will be available at \url{https://github.com/breezexian/UniPhysGen}.
\end{abstract}
    
\section{Introduction}
Recent advances in generative 3D modeling~\cite{xiang2025structured, xiang2025native, zhao2025hunyuan3d, lai2025hunyuan3d} and the emergence of large-scale 3D repositories such as Objaverse~\cite{deitke2023objaverse, deitke2023objaverse_xl} have dramatically increased the availability of high-quality 3D assets. However, despite their realistic geometry and appearance, most existing assets remain fundamentally \emph{simulation-unaware}, lacking physically grounded semantics required for realistic interaction, simulation, and embodied AI. Such semantics include both \textbf{articulation semantics}, such as articulated structures and kinematic parameters, and \textbf{intrinsic physical properties}, including material, density, friction, and mass. Consequently, a gap remains between visually realistic 3D assets and simulation-ready assets.

A central challenge therefore lies in how to automatically transform raw 3D meshes into physically consistent and simulation-ready assets with unified physical semantics. Existing physical grounding approaches remain fragmented, primarily treating articulation grounding and intrinsic physical property estimation as independent problems~\cite{chen2024urdformer, zhao2025real2code, le2025articulate, zhai2024physical, xu2025gaussianproperty}. However, simulation-ready assets require these semantics to be jointly consistent during physical interaction and simulation. Grounding these semantics independently often leads to physically inconsistent motions, unstable interactions, or implausible simulation behaviors. 
More recent physics-grounded generation frameworks~\cite{cao2025physx, NEURIPS2025_86beeac1, yang2026physforge} have further explored jointly modeling articulation and intrinsic physical properties. Nevertheless, these approaches typically rely on canonicalized part decompositions, clean object structures, or structural assumptions constructed under constrained synthetic settings. Such assumptions generalize poorly to the heterogeneous and inconsistent decompositions commonly observed in diverse resources, including artist-designed meshes, procedurally generated structures, and generative 3D assets. 
\textbf{Meanwhile,} as summarized in Table~\ref{tab:dataset_comparison}, existing physically grounded datasets either focus primarily on articulation semantics~\cite{xiang2020sapien, iliash2026s2o, geng2023gapartnet, li2026particulate} or rely heavily on manual curation, clean pre-segmented object parts, and canonical structural decompositions~\cite{NEURIPS2025_86beeac1, yang2026physforge}.
More importantly, although physical grounding aims to enable simulation-ready assets, existing datasets typically stop at semantic annotation, with limited use of simulation to validate or refine the grounded results.


To address these limitations, we introduce \textbf{UniPhys}, an automated physical grounding pipeline that uniformly transforms 3D meshes from diverse sources into simulation-ready assets with unified articulation and physical property semantics. 
As illustrated in Fig.~\ref{fig:uniphys}, UniPhys targets three key challenges in physical grounding, which are addressed through dedicated modules in our system.
\textbf{(1) Physically meaningful structural decomposition.}
Physical grounding requires part decompositions that preserve motion-relevant and material-sensitive structures. Existing methods primarily optimize geometric or semantic consistency~\cite{liu2025partfield, ma2025p3}, often overlooking physically meaningful part abstractions. We therefore introduce a perceptually guided decomposition framework for physical grounding.
\textbf{(2) Geometry-aware articulation grounding.}
Large multimodal models provide strong semantic priors for object functionality, but often struggle to ground these priors into geometrically valid 3D articulation parameters. 
We therefore develop a geometry-aware articulation grounding strategy that combines multimodal semantic understanding with geometric motion reasoning to infer physically consistent articulation configurations.
\textbf{(3) Simulation-driven consistency verification.}
Automatically grounded physical annotations may still contain physically implausible properties or unstable articulated behaviors. To alleviate this issue, we introduce a simulation-driven verification framework that identifies and filters inconsistent annotations through articulated motion consistency and intrinsic physical plausibility checks.


Driven by the UniPhys pipeline, we further construct \textbf{UniPhys-40K}, a large-scale physically grounded dataset containing over 40K 3D assets with unified articulation and intrinsic physical properties aggregated from heterogeneous 3D asset sources, including synthetic assets, CAD models, and artist-designed collections~\cite{mo2019partnet, deitke2023objaverse, khanna2024habitat, collins2022abo, fu20213d}. 
To support rigorous evaluation, we additionally introduce \textbf{UniPhys-Bench}, a high-quality benchmark containing 1,927 articulated objects with 5,469 motion-relevant components, constructed through a labor-intensive human-in-the-loop annotation process. 
Manual verification and correction of both articulation semantics and intrinsic physical properties are performed to ensure physically consistent annotations across heterogeneous 3D assets.

Building upon UniPhys-40K, we propose \textbf{UniPhysGen}, a unified physical grounding model designed for robust reasoning under heterogeneous part decompositions. UniPhysGen jointly models articulation semantics and intrinsic physical properties within a shared representation space without assuming canonicalized object structures. 
To mitigate geometry shortcut learning, UniPhysGen introduces geometry-robust articulation grounding through SO(3) augmentation and spherical articulation axis parameterization. Combined with unified global positional encoding across part-level and object-level geometries, the model achieves stable articulation grounding under diverse structural variations. 


Our main contributions are summarized as follows:

\begin{itemize}[nosep,leftmargin=*]

\item We introduce \textbf{UniPhys}, a scalable pipeline that automatically grounds raw 3D assets with articulation semantics and intrinsic physical properties.

\item We construct \textbf{UniPhys-40K} and \textbf{UniPhys-Bench}, enabling large-scale training and evaluation for physical grounding.

\item We propose \textbf{UniPhysGen}, a unified physical grounding model that jointly reasons over articulation semantics and intrinsic physical properties under heterogeneous part decompositions through geometry-robust articulation grounding and physical semantic alignment.


\item 
Extensive experiments demonstrate that UniPhysGen achieves state-of-the-art performance on articulation and intrinsic physical property grounding across most evaluation settings while remaining robust to heterogeneous part abstractions. We further validate grounded-asset deployability in robotic simulation environments.

\end{itemize}

\begin{table*}[t]
\centering
\small
\setlength{\tabcolsep}{3.2pt}
\renewcommand{\arraystretch}{1.08}
\caption{
Comparison of 3D datasets for physical grounding.
\textbf{S1--S3} denote no pre-segmented parts, low human effort, and raw-asset extensibility.
\cmarkb/\xmarkb: yes/no; --: not applicable or not reported. 
}
\label{tab:dataset_comparison}
\resizebox{\textwidth}{!}{
\begin{tabular}{lcccccccccc}
\toprule
\multirow{2}{*}{\textbf{Dataset}} &
\multirow{2}{*}{\textbf{Primary Data Source}} &
\multirow{2}{*}{\textbf{\#Obj.}} &
\multirow{2}{*}{\textbf{\#Cat.}} &
\multirow{2}{*}{\textbf{\makecell[c]{Articulation Semantics}}} &
\multirow{2}{*}{\textbf{\makecell[c]{Intrinsic Physical \\ Properties}}} &
\multirow{2}{*}{\textbf{\makecell[c]{Simulation \\ verification}}} &
\multicolumn{3}{c}{\textbf{Scalability}} \\
\cmidrule(lr){8-10}
&
&
&
&
&
&
&
\textbf{S1} & \textbf{S2} & \textbf{S3} \\
\midrule

ShapeNet~\cite{chang2015shapenet}
& Online & 51K & 55 
& \xmarkb & \xmarkb & \xmarkb
& -- & -- & -- \\

PartNet~\cite{mo2019partnet}
& ShapeNet & 26K & 24 
& \xmarkb & \xmarkb  & \xmarkb
& -- & -- & -- \\

3D-FUTURE~\cite{fu20213d}
& Industry & 10K & 8 
& \xmarkb & \xmarkb & \xmarkb
& -- & -- & -- \\

ABO~\cite{collins2022abo}
& Amazon & 8K & 63 
& \xmarkb & Partial & \xmarkb
& -- & -- & -- \\

HSSD~\cite{khanna2024habitat}
& Floorplanner & 18K & 466 
& \xmarkb & \xmarkb & \xmarkb
& -- & -- & -- \\

Objaverse~\cite{deitke2023objaverse}
& Multi. & 818K & 21K 
& \xmarkb & \xmarkb & \xmarkb
& -- & -- & -- \\

\midrule

PartNet-Mobility~\cite{xiang2020sapien}
& PartNet & 2.3K & 46 
& \cmarkb & \xmarkb & \xmarkb
& -- & -- & -- \\

GAPartNet~\cite{geng2023gapartnet}
& PartNet & 1.1K & 27 
& \cmarkb & \xmarkb & \xmarkb
& -- & -- & -- \\

PhysXNet~\cite{NEURIPS2025_86beeac1}
& PartNet & 26K & 24 
& \cmarkb & \cmarkb & \xmarkb
& \xmarkb & \xmarkb & \xmarkb \\

PhysDB~\cite{yang2026physforge}
& Objaverse & 150K & 7 
& Joint type only & \cmarkb & \xmarkb
& \xmarkb & \xmarkb & \xmarkb \\

\midrule

\textbf{UniPhys-40K (Ours)}
& \textbf{Multi.}
& \textbf{40K}
& \textbf{84}
& \cmarkb
& \cmarkb
& \cmarkb
& \cmarkb
& \cmarkb
& \cmarkb \\

\bottomrule
\end{tabular}
}
\vspace{-10pt}
\end{table*}
\section{Related Work}

\noindent \textbf{Physical Grounding for 3D Assets.}
Existing physical grounding methods mainly focus on either articulation semantics~\cite{chen2024urdformer,qiu2025articulate,zhao2025real2code,le2025articulate,li2026particulate} or intrinsic physical properties~\cite{zhai2024physical,xu2025gaussianproperty}. Prior work on articulation semantics includes image-based URDF prediction~\cite{chen2024urdformer}, program synthesis~\cite{zhao2025real2code}, and geometry-grounded articulation reasoning from 3D point clouds~\cite{li2026particulate}. In parallel, work on intrinsic physical properties focuses on estimating attributes such as density, friction, and mass through multimodal physical reasoning~\cite{zhai2024physical,xu2025gaussianproperty}. Recent frameworks move toward unified physical grounding by jointly modeling articulation semantics and intrinsic physical properties~\cite{cao2025physx,NEURIPS2025_86beeac1,yang2026physforge}. However, these methods typically assume clean, canonicalized part decompositions, which rarely hold for heterogeneous assets with inconsistent granularity and structural ambiguities.

\noindent \textbf{Physical Grounding Datasets.}
Large-scale repositories such as ShapeNet~\cite{chang2015shapenet}, ABO~\cite{collins2022abo}, and Objaverse~\cite{deitke2023objaverse,deitke2023objaverse_xl} provide abundant geometric and visual content but largely lack physical semantics required for simulation-ready interaction. Subsequent datasets introduce articulation annotations, including PartNet-Mobility~\cite{xiang2020sapien} and GAPartNet~\cite{geng2023gapartnet}. More recent efforts further incorporate intrinsic physical properties alongside articulation semantics~\cite{NEURIPS2025_86beeac1,yang2026physforge}. 
However, these datasets remain largely built on clean, pre-structured assets and often require substantial manual annotation, limiting scalability to heterogeneous 3D assets.

\section{UniPhys: Scalable Unified Physical Grounding Pipeline}
\label{sec:uniphys}

\begin{figure}[htbp]
  \includegraphics[width=\linewidth]{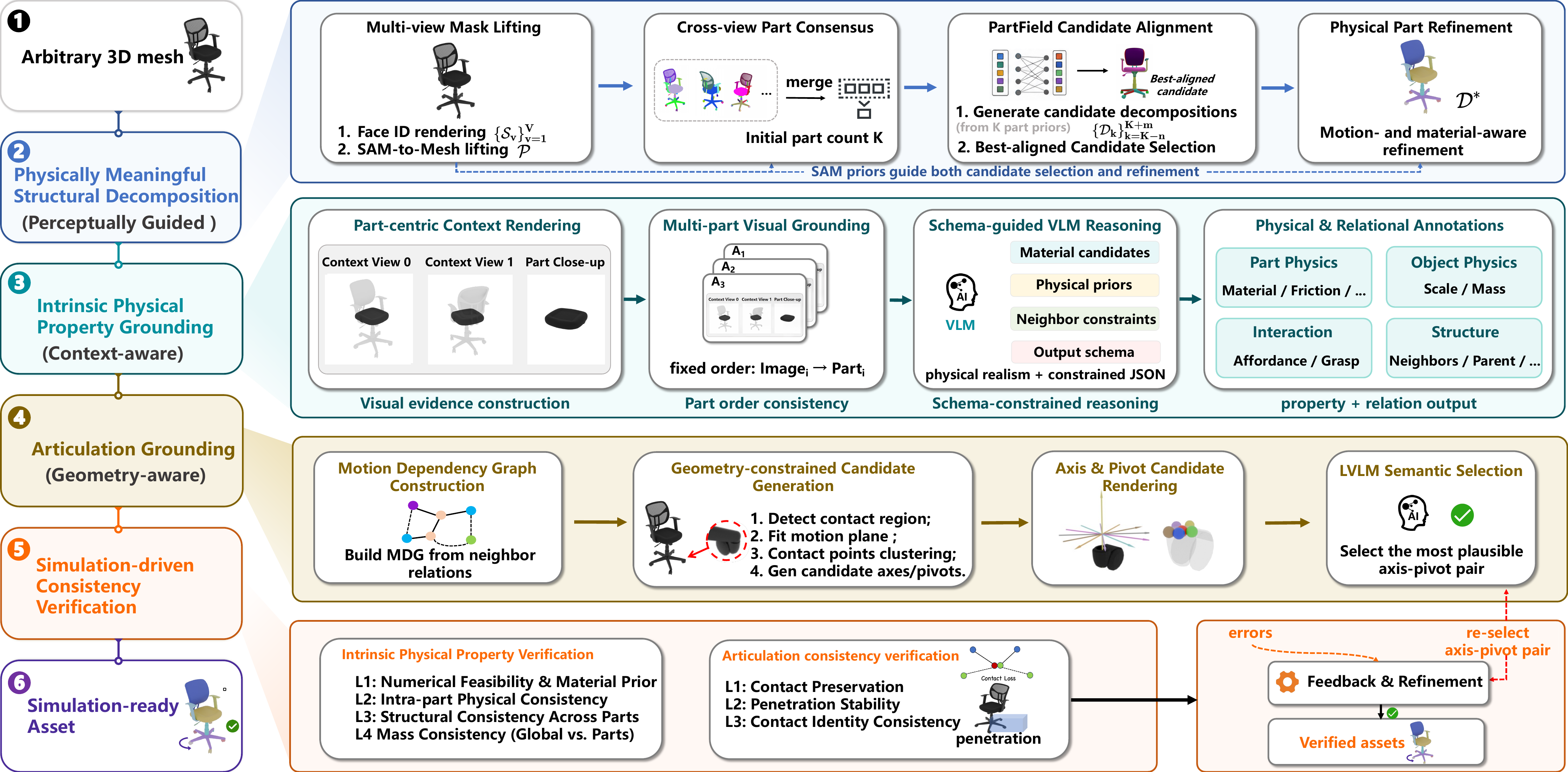}
  \setlength{\abovecaptionskip}{-6pt}
  \caption{UniPhys: A scalable pipeline for transforming heterogeneous 3D assets into simulation-ready assets with unified articulation semantics and intrinsic physical properties.}

  \label{fig:uniphys}
  \vspace{-18pt}
\end{figure}

UniPhys aims to automatically transform heterogeneous 3D assets into simulation-ready assets with grounded articulation and intrinsic physical properties. As shown in Fig.~\ref{fig:uniphys}, the pipeline contains four stages: physically meaningful part decomposition, intrinsic physical property grounding, geometry-aware articulation grounding, and simulation-driven consistency verification. The pipeline is detailed in Sec.~\ref{sec:uniphys_1}. Scaling this pipeline over large 3D asset repositories yields UniPhys-40K and the manually verified UniPhys-Bench benchmark, introduced in Sec.~\ref{sec:uniphys_2}.

\subsection{Uniphys pipeline}
\label{sec:uniphys_1}
\noindent \textbf{Physically Meaningful Structural Decomposition.} 
Existing decomposition methods~\cite{ma2025p3, liu2025partfield, yan2025x}
optimize geometric or semantic consistency, whereas simulation-ready grounding additionally requires articulation, material, and functional coherence.
Our key insight is that perceptual grouping priors learned by segmentation foundation models often coincide with physically meaningful structures, providing effective guidance for physical decomposition.
Based on this observation, UniPhys first lifts multi-view SAM~\cite{ravi2025sam} segmentations into mesh space to build perceptual priors $\mathcal{P}$ and infer a part-cardinality prior $K$. Guided by $K$, PartField~\cite{liu2025partfield} generates candidate decompositions $\{\mathcal{D}_k\}_{k=K-n}^{K+m}$ at varying granularities, which are then aligned to $\mathcal{P}$ through Hungarian matching~\cite{kuhn1955hungarian}. 
Finally, treating components associated with each lifted SAM mask in the best-aligned decomposition as transactions, we use Apriori-based frequent itemset mining~\cite{apriori} to identify stable cross-view component groupings and refine the final decomposition $\mathcal{D}^*$.


\noindent \textbf{Intrinsic Physical Property Grounding.}
Given the physically grounded part decomposition, UniPhys infers intrinsic physical properties through part-aware multimodal reasoning. We construct context-preserving part-centric observations and organize them into a unified part atlas for schema-guided LVLM reasoning. Conditioned on object context, neighboring relations, and physical priors, the LVLM jointly predicts part- and object-level physical properties. 

\noindent \textbf{Geometry-aware Articulation Grounding.}
While LVLMs provide strong semantic priors for understanding object functionality and affordances, directly predicting articulation parameters in continuous 3D space remains geometrically unstable. UniPhys instead constructs feasible articulation hypotheses from neighboring relations and local contact geometry, reducing articulation grounding to semantic selection over geometry-constrained candidates. This combination of geometric constraints and semantic reasoning improves both robustness and physical validity. 

\noindent \textbf{Simulation-driven Consistency Verification.}
Automatically grounded annotations may still be physically inconsistent. Rather than recovering exact real-world physical quantities, UniPhys evaluates whether grounded semantics satisfy simulation-consistent interaction constraints through a simulation-driven verification stage. Invalid annotations are subsequently filtered or refined through lightweight feedback. 
Detailed implementations of all four stages are deferred to Appendix~\ref{sec:appendix_uniphys}.

\subsection{UniPhys-40K and Benchmark.}
\label{sec:uniphys_2}
\begin{figure}[t]
  \includegraphics[width=\linewidth]{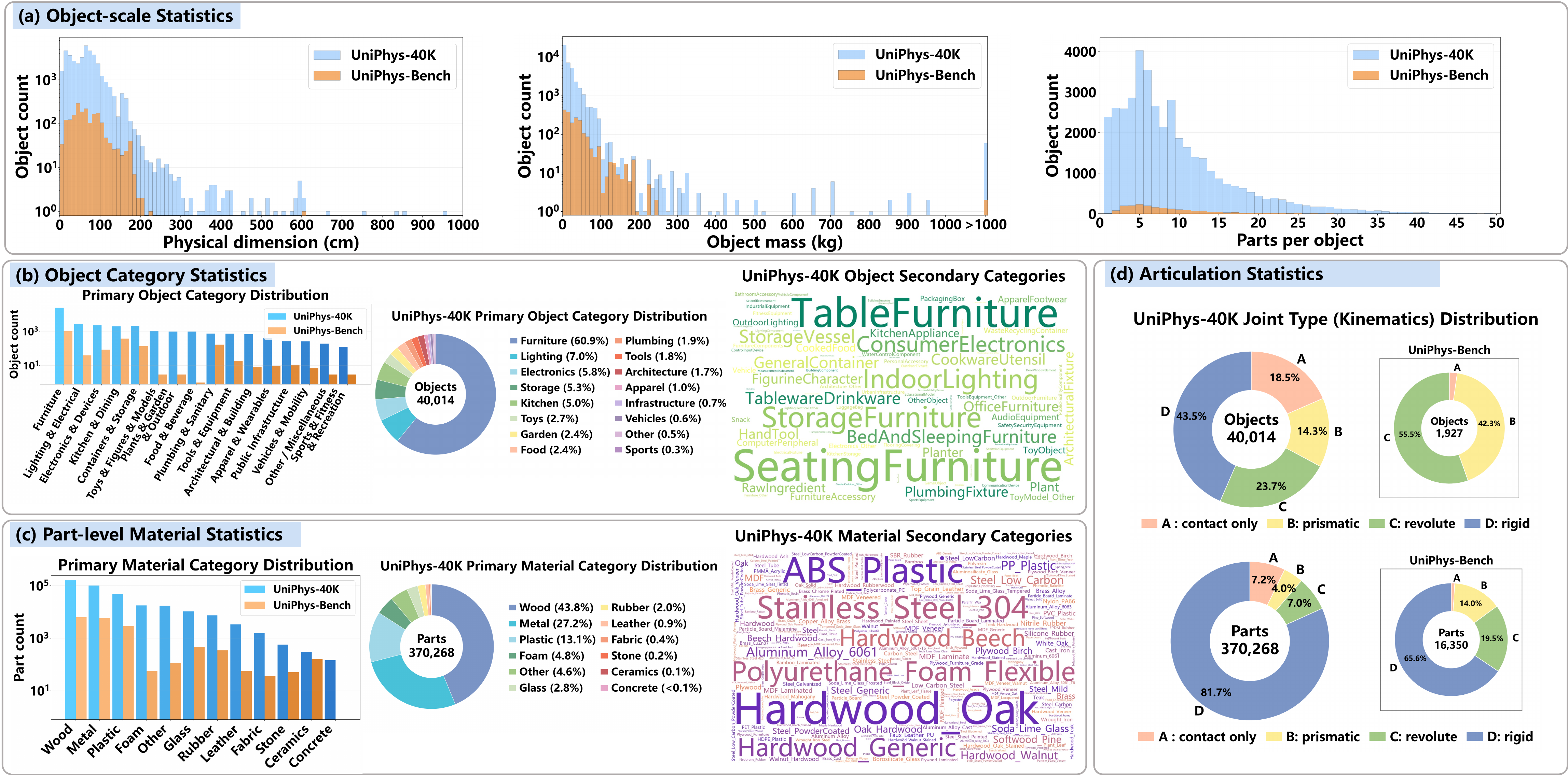}
  \setlength{\abovecaptionskip}{-6pt}
  \caption{ Statistics of UniPhys-40K and UniPhys-Bench.
  }

  \label{fig:statistics}
  \vspace{-10pt}
\end{figure}

\noindent \textbf{Dataset Overview.}
Driven by the proposed UniPhys pipeline, we construct UniPhys-40K, 
\begin{wrapfigure}{r}{0.45\textwidth}
\vspace{-2pt}
\centering
\captionsetup{font=normalsize}
\setlength{\abovecaptionskip}{2pt}
\setlength{\belowcaptionskip}{-2pt}
\includegraphics[width=\linewidth]{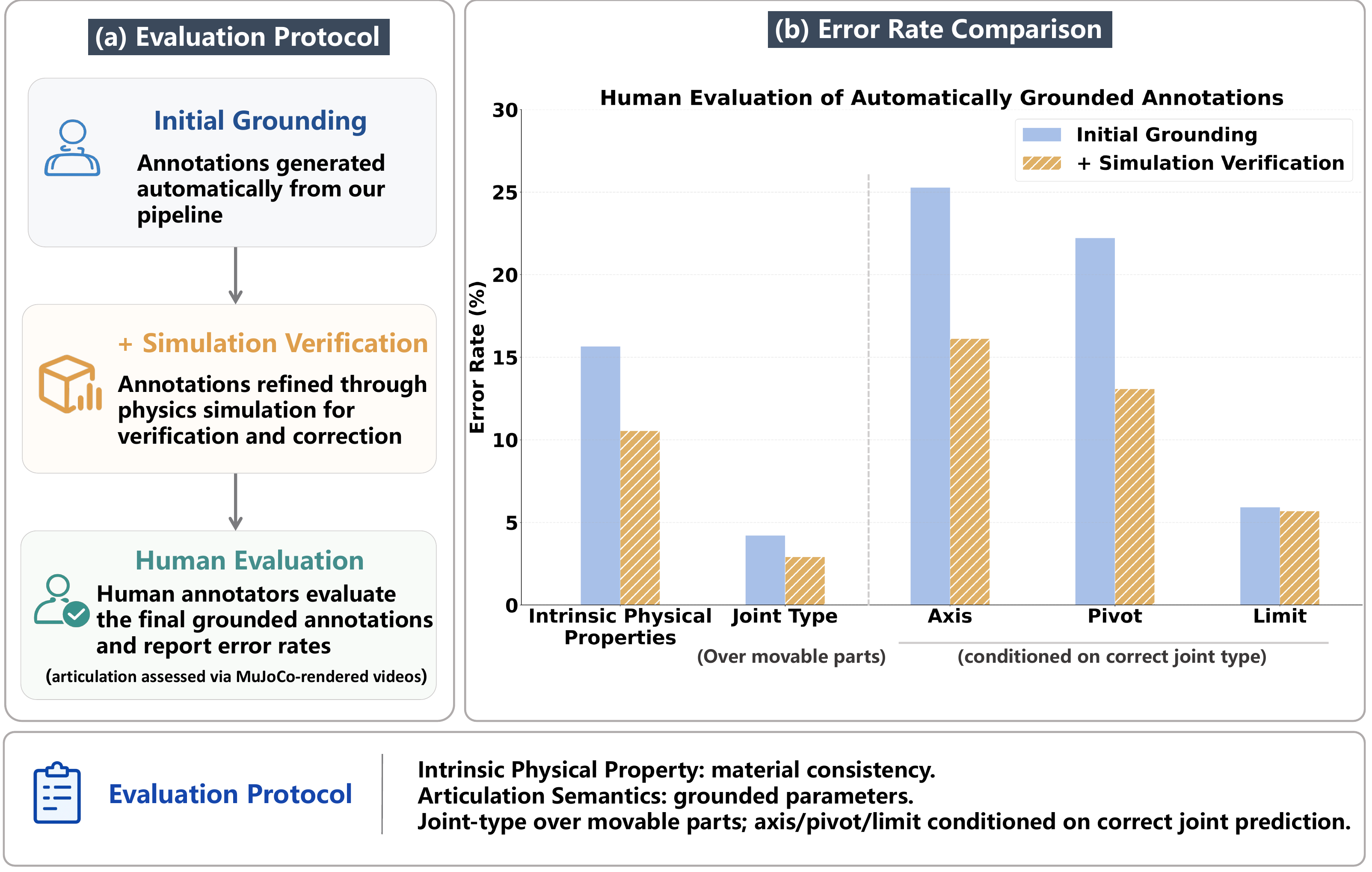}
\caption{Human evaluation.
}
\label{fig:verification}
\vspace{-10pt}
\end{wrapfigure}
a large-scale dataset with grounded articulation semantics and intrinsic physical properties.
UniPhys-40K is built from diverse 3D repositories, including Objaverse~\cite{deitke2023objaverse}, HSSD~\cite{khanna2024habitat}, 3D-FUTURE~\cite{fu20213d}, ABO~\cite{collins2022abo}, and PartNet~\cite{mo2019partnet}. 
To support rigorous evaluation, we further construct UniPhys-Bench, a carefully verified benchmark containing 1,927 articulated objects with 5,469 motion-relevant components spanning diverse object categories and structural complexities. 
As summarized in Fig.~\ref{fig:statistics}, 
UniPhys-40K covers a broad spectrum of object scales, masses, categories, materials, and articulation patterns, ranging from small household objects to large furniture and infrastructure assets. 
The dataset further exhibits long-tailed distributions of both physical properties and kinematic structures, covering rigid, revolute, prismatic, and contact-only interaction relationships.
More dataset construction details are provided in Appendix~\ref{sec:appendix_uniphys_5}.

\noindent \textbf{Quality Analysis.}
We further perform human evaluation on sampled assets before and after simulation-driven verification. As shown in Fig.~\ref{fig:verification}, the results show that verification substantially improves the reliability and physical consistency of automatically grounded simulation-ready assets. This observation supports the effectiveness of simulation-driven consistency verification as a scalable quality-control mechanism for UniPhys.

\section{UniPhysGen}
\label{sec:uniphysgen}

\begin{figure}[htbp]
  \includegraphics[width=\linewidth]{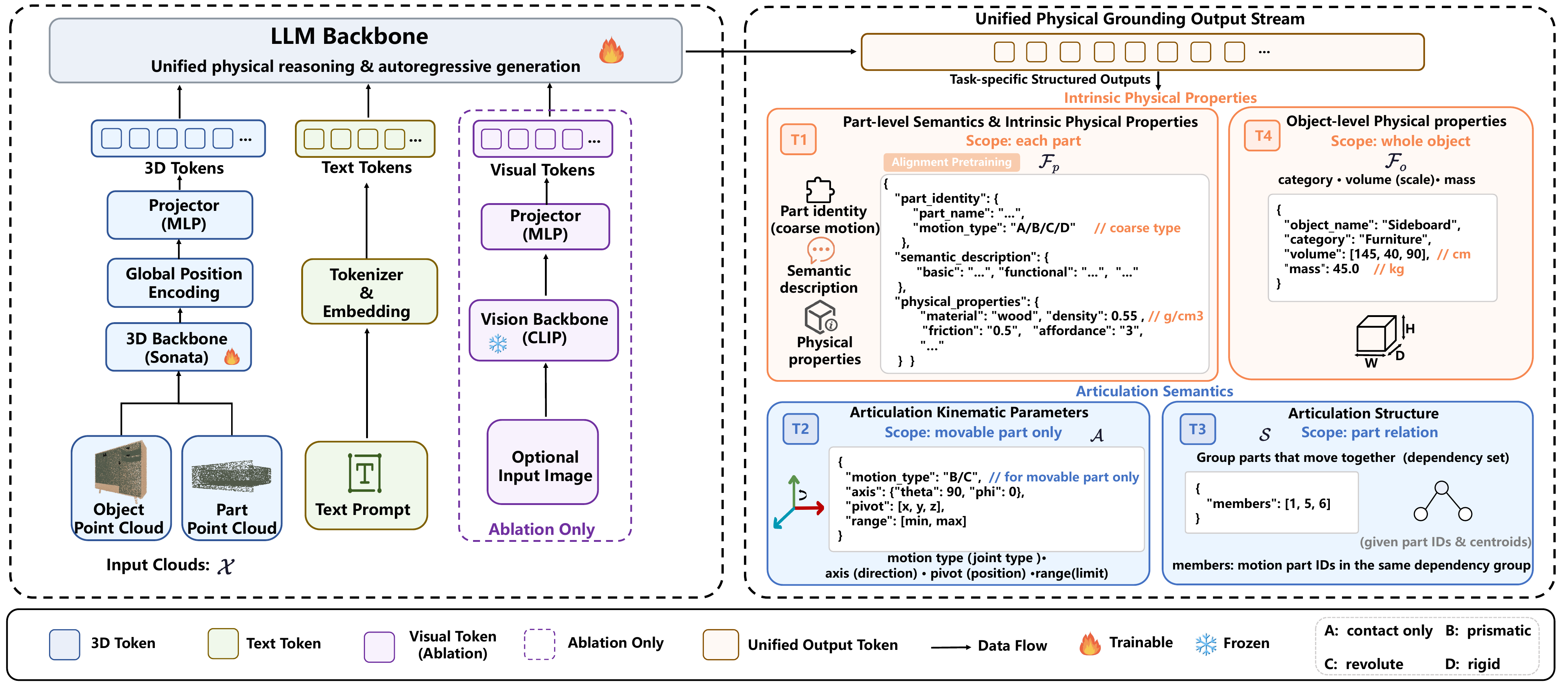}
  \setlength{\abovecaptionskip}{-6pt}
  \caption{Overview of UniPhysGen. The model takes 3D object and part geometry as input (Left) and generates four types of structured physical semantics (Right): part-level properties, articulation kinematics, articulation structure, and object-level properties via a unified multimodal backbone.}
  \label{fig:uniphysgen}
  \vspace{-10pt}
\end{figure}

We formalize UniPhysGen as learning a unified physical grounding framework that maps input 3D geometry to structured physical semantics. Given an object mesh and a target part mesh, we convert them into point clouds and denote the resulting input as $\mathcal{X}$. UniPhysGen predicts four categories of physical semantics:
\begin{equation}
\mathcal{Y}=f_\theta(\mathcal{X}),
\qquad
\mathcal{Y}=\{\mathcal{F}_p,\mathcal{A},\mathcal{S},\mathcal{F}_o\},
\end{equation}
where $f_{\theta}$ denotes a shared UniPhysGen architecture instantiated across all four tasks. Besides, $\mathcal{F}_p$, $\mathcal{A}$, $\mathcal{S}$, and $\mathcal{F}_o$ denote part-level intrinsic physical properties, articulation kinematics, articulation structure, and object-level intrinsic physical properties, respectively.
Specifically, UniPhysGen adopts Qwen3~\cite{yang2025qwen3} as the backbone and performs physical grounding through four stages shown in Fig.~\ref{fig:uniphysgen}: (1) physical semantic alignment pretraining for $\mathcal{F}_p$; (2) geometry-robust articulation grounding for $\mathcal{A}$, including joint type, axis, pivot, and motion limits, on motion-related parts obtained either from the physical semantic alignment stage or from external annotations; (3) articulation structure grounding for $\mathcal{S}$; and (4) object-level physical grounding for $\mathcal{F}_o$.
Unlike prior approaches that rely on canonicalized structures or fixed decomposition templates, UniPhysGen directly operates on heterogeneous 3D asset decompositions. 
Additional details of the four stages are provided in Appendix~\ref{sec:appendix_uniphysgen_1}--\ref{sec:appendix_uniphysgen_4}.

\subsection{Physical Semantic Alignment Pretraining}
A key challenge of unified physical grounding is learning a shared representation space that jointly captures geometry, articulation semantics, and intrinsic physical properties. To this end, we introduce a physical semantic alignment pretraining stage that aligns local part geometry and global object context with structured physical semantic descriptions, including part identity, functional semantics, articulation behavior, and intrinsic physical properties. This representation serves as a unified grounding space shared across downstream tasks, enabling robust physical reasoning under heterogeneous part decompositions. 


\subsection{Geometry-Robust Articulation Grounding}
\label{sec:uniphysgen_2}
\begin{wrapfigure}{r}{0.5\textwidth}
\vspace{-8pt}
\centering
\captionsetup{font=normalsize}
\setlength{\abovecaptionskip}{2pt}
\setlength{\belowcaptionskip}{-1pt}
\includegraphics[width=\linewidth]{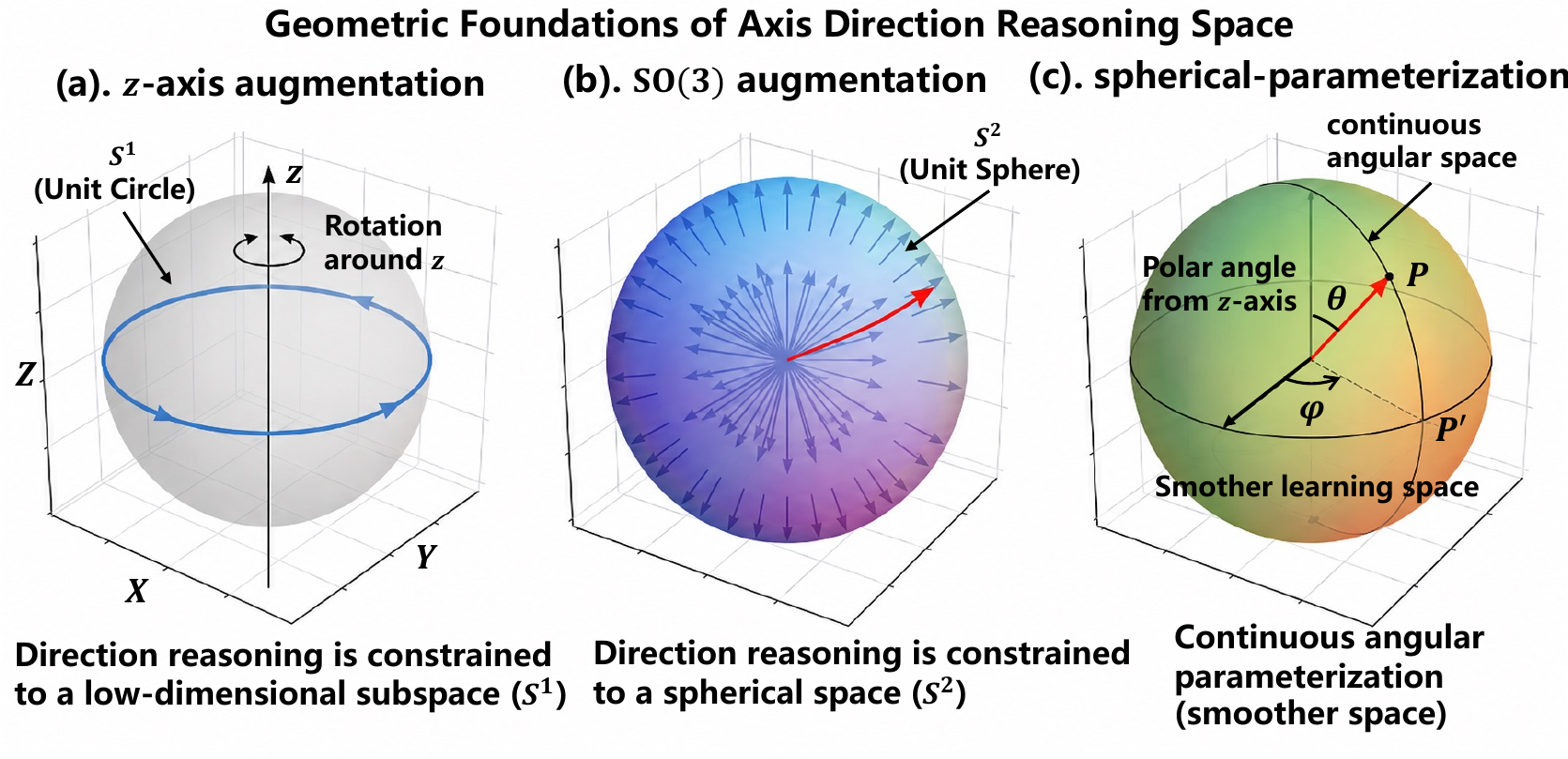}
\caption{Axis direction reasoning space.}
\label{fig:axis_aug}
\vspace{-8pt}
\end{wrapfigure}

Articulation grounding is highly sensitive to geometric shortcuts and rotational bias. 
Existing methods often assume upright object orientations~\cite{li2026particulate} and apply z-axis rotation augmentation, 
$
\mathbf{p}' = \mathbf{R}_{z}(\theta)\mathbf{p},
$
where $\mathbf{p}\in\mathbb{R}^3$ is a 3D point coordinate and $\mathbf{R}_{z}(\theta)$ denotes rotation around the vertical axis (Fig.~\ref{fig:axis_aug}(a)). This encourages view-dependent motion reasoning and generalizes poorly to geometrically incomplete or structurally ambiguous assets.

To mitigate this issue, we replace canonical-orientation augmentation with SO(3)-based rotation augmentation (Fig.~\ref{fig:axis_aug}(b)): 
$
\mathbf{p}' = \mathbf{R}\mathbf{p}, \quad \mathbf{R}\in SO(3) $.
This significantly improves structure-aware articulation reasoning by forcing the model to rely on motion-relevant geometry semantics rather than view-dependent shortcuts. 

However, we observe that SO(3)-based augmentation substantially destabilizes articulation axis reasoning due to directional ambiguity under arbitrary rotations. We therefore represent normalized articulation axes with spherical coordinates instead of directly generating Cartesian directions (Fig.~\ref{fig:axis_aug}(c)):
$
\theta = \arccos(z), \quad
\phi = \operatorname{atan2}(y, x)
$, 
where $\theta$ and $\phi$ denote the polar and azimuth angles, respectively. 
This formulation removes directional sign ambiguity and provides a smoother angular representation space for articulation reasoning. 

To improve pivot estimation, we further introduce a shared global positional encoding for both part-level and object-level point clouds. Specifically, instead of independently normalizing local coordinates, all geometries share the same global voxel coordinate system:
\begin{equation}
\mathbf{g}_i =
\left\lfloor
\frac{\mathbf{p}_i}{s}
\right\rfloor
-
\mathbf{g}_{\min}^{\text{obj}},
\label{eq:global_grid}
\end{equation}
where $\mathbf{g}_{\min}^{\text{obj}}$ is computed from the complete object geometry. The resulting voxel coordinates are subsequently encoded using Fourier positional encoding. Compared with independent local normalization, this strategy preserves global spatial correspondence and substantially improves pivot localization under heterogeneous decompositions.

\subsection{Articulation Structure Grounding}
Building upon the shared physical grounding representation, UniPhysGen predicts motion-dependent part groups that jointly participate in articulated motion. By reasoning over part identities, geometric locations, and global object context, the model captures articulation-aware structural dependencies without relying on canonical hierarchies or fixed decomposition templates. 

\subsection{Object-Level Physical Grounding}
Finally, UniPhysGen performs holistic object-level physical reasoning by predicting global physical properties such as object category, scale, and mass. Sharing the pretrained physical semantic representation across part-level and object-level tasks enables physically consistent reasoning across semantic granularities while maintaining robustness under heterogeneous 3D assets. 
\section{Experiments}

\subsection{Experimental Setup}

\noindent\textbf{Implementation Details.}
UniPhysGen is built upon Qwen3-1.7B~\cite{yang2025qwen3} with Sonata~\cite{wu2025sonata} as the point cloud encoder. Following SpatialLM~\cite{mao2025spatiallm}, all grounding tasks are trained using full-parameter fine-tuning. For the image-modality ablation experiments, as illustrated in Fig.~\ref{fig:uniphysgen}, we incorporate frozen CLIP ViT-L/14~\cite{radford2021learning} features through a lightweight projector and apply LoRA fine-tuning. More implementation details are in Appendix~\ref{sec:appendix_implementation}.

\noindent\textbf{Benchmarks, Baselines, and Metrics.}
We primarily evaluate unified physical grounding on UniPhys-Bench, which contains jointly annotated articulation semantics and intrinsic physical properties. Additional articulation evaluation on PartNet-Mobility~\cite{xiang2020sapien} is provided in Appendix~\ref{sec:appendix_res_pm}.
For articulation grounding, we compare with Real2Code~\cite{zhao2025real2code}, URDFormer~\cite{chen2024urdformer}, Articulate-Anything~\cite{le2025articulate}, and PARTICULATE~\cite{li2026particulate}. For intrinsic physical property grounding, we compare against NeRF2Physics~\cite{zhai2024physical}.
Kinematic parameters are evaluated only on ground-truth movable parts, decoupling kinematic estimation from movable-part identification.
Following prior articulation evaluation protocols~\cite{le2025articulate}, we report joint type accuracy, axis, pivot, and motion limit metrics, together with articulation structure grounding. For intrinsic physical properties, following NeRF2Physics~\cite{zhai2024physical}, we report ALDE and MnRE for mass and scale estimation. Density is evaluated using ALDE, while friction and affordance grounding measured by MAE, and material prediction by classification accuracy. Detailed evaluation protocols, fairness adaptations, and metric definitions are provided in Appendix~\ref{sec:appendix_metrics} and Appendix~\ref{sec:appendix_fairness}.

\begin{table*}[htbp]
\centering
\caption{Quantitative evaluation of unified physical grounding on UniPhys-Bench.}
\label{tab:total_res}
\resizebox{\textwidth}{!}{
\begin{tabular}{@{}lcccc|cc|ccc|cccc|c@{}}
\toprule

\multirow{3}{*}{\makecell[lt]{\\ \textbf{Method}}}

& \multicolumn{6}{c|}{\textbf{Articulation Semantics}}
& \multicolumn{8}{c}{\textbf{Intrinsic Physical Properties}} \\

\cmidrule(lr){2-7}
\cmidrule(lr){8-15}

& \multicolumn{4}{c|}{\textbf{Kinematic Parameters}}
& \multicolumn{2}{c|}{\textbf{Articulation Structure}}

& \multicolumn{3}{c|}{\textbf{Material Properties}}
& \multicolumn{4}{c|}{\textbf{Scale \& Mass Properties}}
& \multicolumn{1}{c}{\textbf{Affordance}} \\

\cmidrule(lr){2-5}
\cmidrule(lr){6-7}
\cmidrule(lr){8-10}
\cmidrule(lr){11-14}
\cmidrule(lr){15-15}

& Joint Type
& Axis
& Pivot
& Limit

& Struct.
& Struct.

& Mat.
& $\rho$
& $\mu$

& Dim.
& Dim.
& Mass
& Mass

& Aff. \\

& Acc $\uparrow$
& Ang.\ Err ($^\circ$) $\downarrow$
& Dist.\ Err $\downarrow$
& mIoU $\uparrow$

& mIoU $\uparrow$
& F1 $\uparrow$

& Acc $\uparrow$
& ALDE $\downarrow$
& MAE $\downarrow$

& ALDE $\downarrow$
& MnRE $\uparrow$
& ALDE $\downarrow$
& MnRE $\uparrow$

& MAE $\downarrow$ \\

\midrule

Real2Code & 60.13 & 52.26 & 0.317 & -- & -- & -- & -- & -- & -- & -- & -- & -- & -- & -- \\
URDFormer & 62.13 & 10.78 & 0.427 & 79.61 & -- & -- & -- & -- & -- & -- & -- & -- & -- & -- \\
Articulate-Anything$^{*}$ & \textbf{94.09} & \textbf{5.65} & 0.108 & 72.16 & -- & -- & -- & -- & -- & -- & -- & -- & -- & -- \\
PARTICULATE & 77.17 & 12.80 & 0.261 & 63.42 & -- & -- & -- & -- & -- & -- & -- & -- & -- & -- \\
NeRF2Physics & -- & -- & -- & -- & -- & -- & 53.09 & 0.807 & 0.142 & -- & -- & 0.788 & 0.524 & -- \\
\midrule

UniPhysGen (Ours)
& 89.96 & 9.77 & \textbf{0.099} & \textbf{84.95}
& \textbf{80.46} & \textbf{83.21}

& \textbf{77.92} & \textbf{0.438} & \textbf{0.062}
& \textbf{0.251} & \textbf{0.819} & \textbf{0.664} & \textbf{0.663}
& \textbf{1.951} \\

\bottomrule
\end{tabular}
}
\footnotesize
\textit{Note: $\rho$ and $\mu$ denote density and friction coefficient, respectively. 
* Articulate-Anything relies on semantic priors, canonical coordinates, and in-context articulation templates.}
\vspace{-10pt}
\end{table*}

\begin{figure}[htbp]
  \includegraphics[width=\linewidth]{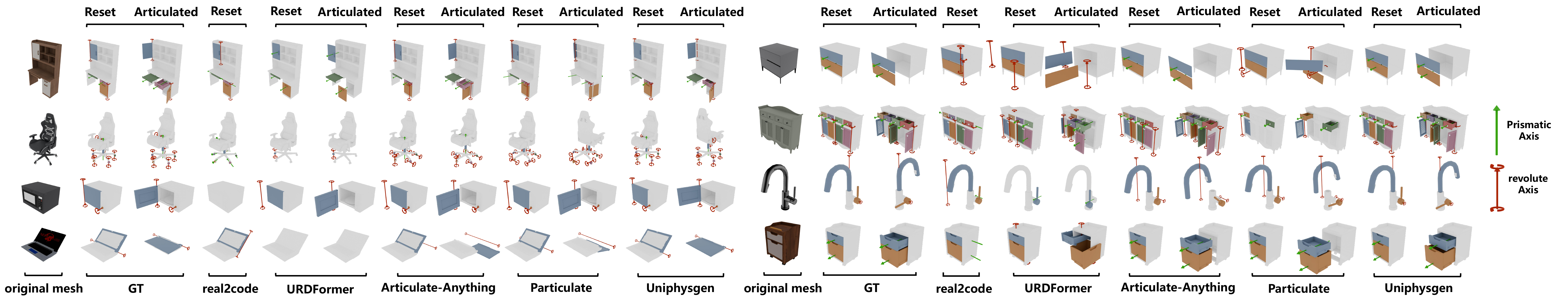}
  \setlength{\abovecaptionskip}{-6pt}
  \caption{Qualitative comparison of articulation grounding on UniPhys-Bench. “Reset” denotes the initial object state, while “Articulated” denotes the maximally articulated state.
  }
  \label{fig:qualitative_compare}
  \vspace{-10pt}
\end{figure}

\begin{figure}[htbp]
  \includegraphics[width=\linewidth]{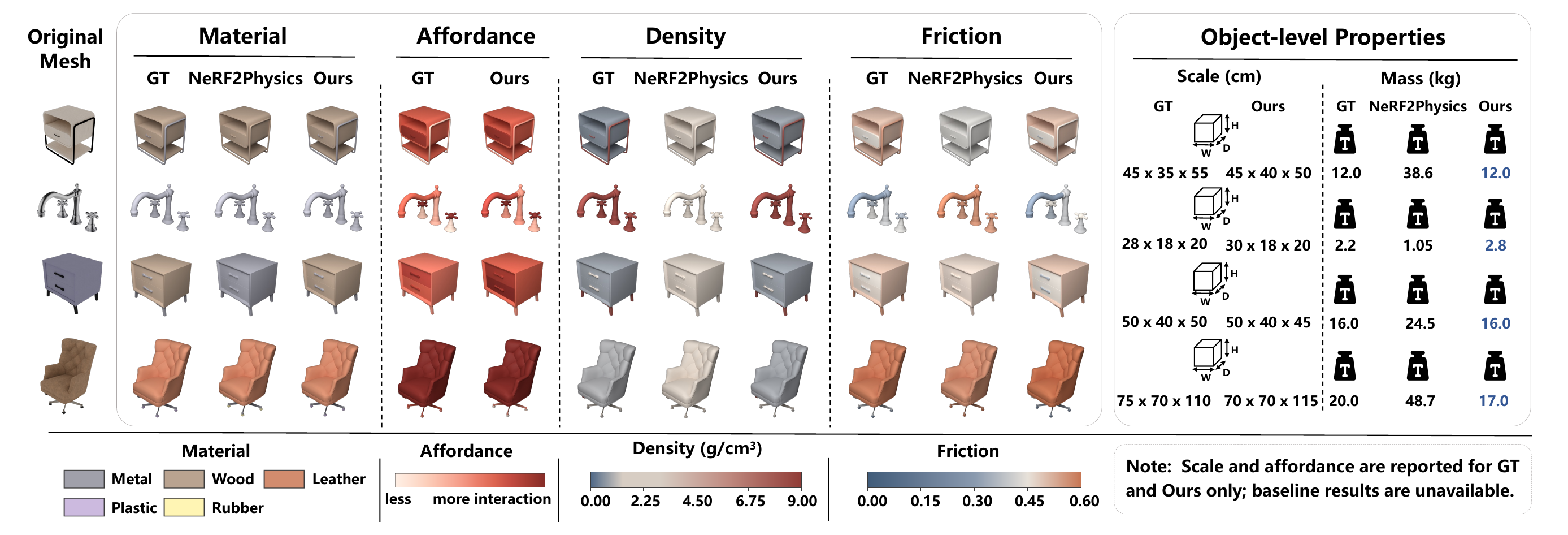}
  \setlength{\abovecaptionskip}{-6pt}
  \caption{Qualitative comparison of intrinsic physical property grounding on UniPhys-Bench.}
  \label{fig:qualitative_compare_basic}
  \vspace{-10pt}
\end{figure}

\subsection{Main results}
\noindent \textbf{Unified Physical Grounding Results.} 
Table~\ref{tab:total_res} reports quantitative results on unified physical grounding, covering articulation semantics and intrinsic physical properties. UniPhysGen consistently outperforms prior methods across most metrics, with clear advantages in articulation structure reasoning, pivot localization, motion limit estimation, and intrinsic physical property grounding, demonstrating the effectiveness of unified physical semantic reasoning.
Fig.~\ref{fig:qualitative_compare} further shows that geometry-centric methods such as PARTICULATE are susceptible to geometric shortcuts on incomplete or ambiguous assets, while Articulate-Anything often produces physically inaccurate pivots and motion limits despite competitive joint-type prediction.
In contrast, UniPhysGen achieves more stable articulation grounding by jointly leveraging geometric and physical semantic cues.
Beyond articulation grounding, Fig.~\ref{fig:qualitative_compare_basic} further shows that UniPhysGen produces more coherent material, affordance, scale, and mass estimates.
More results are provided in Appendix~\ref{sec:appendix_res}


\begin{table*}[htbp]
\centering
\footnotesize
\setlength{\tabcolsep}{5pt}
\caption{
Articulation generalization under different rotation augmentation and axis parameterization.
}
\label{tab:rotation_ablation}
\resizebox{\textwidth}{!}{
\begin{tabular}{lccccccc}
\toprule

\multirow{2}{*}{\textbf{Training Strategy}}
& \multirow{2}{*}{\textbf{Test Condition}}

& \multicolumn{5}{c}{\textbf{Motion Parameters}} \\

\cmidrule(lr){3-7}

&

& Joint Type
& Axis
& Axis
& Pivot
& Limit \\

&

& Acc $\uparrow$
& Ang.\ Err ($^\circ$) $\downarrow$
& Acc $\uparrow$
& Dist.\ Err $\downarrow$
& mIoU $\uparrow$ \\

\midrule

\multirow{3}{*}{\makecell[l]{Canonical Rotation\\(Z-axis)}}

& Canonical-orientation
& 80.63 & 9.46 & 89.45 & 0.107 & 84.51 \\

& Rotated-Z
& 81.67 & 9.68 & 88.81 & 0.097 & 84.20 \\

& Challenging-Rotated
& 50.09 & 13.19 & 85.27 & 0.075 & 80.96 \\

\midrule

\multirow{3}{*}{\makecell[l]{SO(3)-based Rotation}}

& Canonical-orientation
& 90.12 & 6.28 & 93.16 & 0.096 & 85.09 \\

& Rotated-Z
& 86.39 & 19.39 & 69.96 & 0.118 & 82.94 \\

& Challenging-Rotated
& 80.53 & 30.61 & 55.46 & 0.081 & 79.25 \\

\midrule

\multirow{3}{*}{\makecell[l]{SO(3)-based Rotation\\+ Spherical Augmentation}}

& Canonical-orientation
& 89.96 & 9.77 & 89.33 & 0.099 & 84.95 \\

& Rotated-Z
& 86.25 & 15.29 & 82.42 & 0.105 & 83.82 \\

& Challenging-Rotated
& 78.11 & 23.40 & 74.05 & 0.085 & 80.42 \\

\bottomrule
\end{tabular}
}
\textit{
Note: Axis Acc denotes the percentage of predicted axes within $10^\circ$ of ground truth. Canonical-Orientation uses original orientations, Rotated-Z applies random z-axis rotations, and Challenging-Rotated evaluates the challenging subset under the same rotations.
}
\vspace{-8pt}
\end{table*}

\begin{table}[htbp]
\centering
\footnotesize
\setlength{\tabcolsep}{5pt}

\begin{minipage}[t]{0.47\linewidth}
\centering
\caption{
Robustness analysis under different part composition granularities.
}
\label{tab:granularity}
\resizebox{\linewidth}{!}{%
\begin{tabular}{lcccc}
\toprule

\multirow{3}{*}{\textbf{Part Composition}}
& \multicolumn{4}{c}{\textbf{Motion Parameters}} \\

\cmidrule(lr){2-5}

& Joint Type
& Axis
& Pivot
& Limit \\

& Acc $\uparrow$
& Ang.\ Err ($^\circ$) $\downarrow$
& Dist.\ Err $\downarrow$
& mIoU $\uparrow$ \\

\midrule

Original Parts
& 89.96 & \textbf{9.77} & \textbf{0.099} & 84.95
\\

Merged Parts
& \textbf{90.03} & 9.85 & 0.105 & \textbf{85.04}
\\

\bottomrule
\end{tabular}
}
\end{minipage}
\hfill
\begin{minipage}[t]{0.49\linewidth}
\centering
\caption{
Ablation study on positional encoding strategies for 3D point cloud tokenization.
}
\label{tab:pos_encoder}
\resizebox{\linewidth}{!}{%
\begin{tabular}{lcccc}
\toprule

\multirow{3}{*}{\textbf{Positional Encoding}}
& \multicolumn{4}{c}{\textbf{Motion Parameters}} \\

\cmidrule(lr){2-5}

& Joint Type
& Axis
& Pivot
& Limit \\

& Acc $\uparrow$
& Ang.\ Err ($^\circ$) $\downarrow$
& Dist.\ Err $\downarrow$
& mIoU $\uparrow$ \\

\midrule

Separate PE
& 83.26
& 11.59
& 0.319
& 83.72 \\

Unified Global PE
& \textbf{89.96}
& \textbf{9.77}
& \textbf{0.099}
& \textbf{84.95} \\

\bottomrule
\end{tabular}
}
\end{minipage}
\vspace{-6pt}
\end{table}

\subsection{Robustness and Ablation Analysis}
\noindent \textbf{Robust Articulation Grounding Analysis.}
\textbf{(1) Geometric Shortcut Mitigation.}
As shown in Table~\ref{tab:rotation_ablation}, we compare canonical rotation augmentation, SO(3) augmentation, and spherical axis parameterization under different test conditions. To stress-test structurally incomplete assets, we construct a Challenging articulation subset containing 360 assets and 1,175 articulated parts, including drawer front panels without underlying drawer bodies. 
Under canonical rotation augmentation, UniPhysGen performs well on Canonical-Orientation but degrades substantially on Challenging-Rotated, where isolated drawer panels are frequently misclassified as revolute joints due to geometric shortcut bias.
SO(3) augmentation improves joint type accuracy on Challenging-Rotated from 50.09\% to 80.53\%, but makes Cartesian axis prediction less stable under Rotated-Z as the reasoning space expands from canonical planar directions to full spherical orientations (Fig.~\ref{fig:axis_aug}). By further introducing spherical axis parameterization, UniPhysGen achieves more robust axis estimation while maintaining balanced performance across test conditions.
\textbf{(2) Heterogeneous Part Decomposition Robustness.}
We further evaluate UniPhysGen under heterogeneous part decompositions by merging motion-dependent auxiliary components into primary articulated parts according to annotated dependency relations, resulting in coarser decompositions with over 60\% merged components. As shown in Table~\ref{tab:granularity}, UniPhysGen maintains strong articulation grounding performance across different decomposition granularities. Notably, joint type and motion limit estimation even improve slightly after merging, suggesting robustness to structurally diverse part abstractions.

\noindent \textbf{Ablation Study.}
\textbf{(1) Global Positional Encoding.}
As shown in Table~\ref{tab:pos_encoder}, the proposed unified global positional encoding consistently improves articulation grounding over separate part- and object-level encodings, especially reducing pivot error from 0.319 to 0.099. This demonstrates the benefit of globally consistent spatial indexing for cross-scale geometric reasoning between parts and objects.
\textbf{(2) Image Modality Ablation.}
Table~\ref{tab:img_res} shows that image features improve appearance-related properties such as material classification, density, and friction, but provide limited overall gains. For geometry- and functionality-dominated properties such as scale, mass, and affordance, adding image features leads to slight performance degradation, suggesting that 3D geometry and structural-functional context remain the dominant cues for these properties.
\textbf{(3) Model Scaling Analysis.}
Table~\ref{tab:scale_res} compares different backbone sizes. Scaling from 0.6B to 1.7B consistently improves performance across both articulation semantics and intrinsic physical properties, particularly for structure reasoning and material-aware grounding, indicating that larger models better support unified geometric-semantic reasoning.

\subsection{Simulation-Based Embodied Interaction}
\begin{figure}[htbp]
  \includegraphics[width=\linewidth]{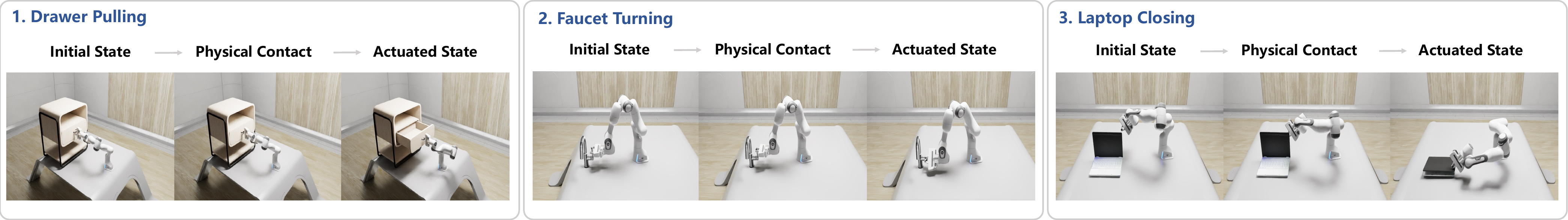}
  \caption{
Qualitative simulation-based validation of UniPhysGen-grounded assets.
Each sequence shows the initial state, physical contact, and the actuated state after robot interaction. 
}
  \label{fig:sim}
    \vspace{-10pt}
\end{figure}

We qualitatively validate UniPhysGen in physics simulation (Isaac Sim) by integrating predicted articulation semantics and intrinsic physical properties into simulation-ready assets and evaluating representative embodied interactions, including drawer pulling, faucet turning, and laptop closing. As shown in Fig.~\ref{fig:sim}, the grounded assets support stable and physically consistent interactions in simulation.

\begin{table*}[htbp]
\centering
\caption{Ablation study on physical property estimation.}
\label{tab:img_res}
\resizebox{0.8\textwidth}{!}{
\begin{tabular}{@{}lccc|cccc|c@{}}
\toprule

\multirow{2}{*}{\textbf{Method}}

& \multicolumn{3}{c|}{\textbf{Material Properties}}
& \multicolumn{4}{c|}{\textbf{Scale \& Mass Properties}}
& \textbf{Affordance} \\

\cmidrule(lr){2-4}
\cmidrule(lr){5-8}
\cmidrule(lr){9-9}

& Mat.
& $\rho$
& $\mu$

& Dim.
& Dim.
& Mass
& Mass

& Aff. \\

& Acc $\uparrow$
& ALDE $\downarrow$
& MAE $\downarrow$

& ALDE $\downarrow$
& MnRE $\uparrow$
& ALDE $\downarrow$
& MnRE $\uparrow$

& MAE $\downarrow$ \\

\midrule

UniPhysGen (w/o Image Modality)

& 77.92 & 0.438 & 0.062
& \textbf{0.251} & \textbf{0.819} & \textbf{0.664} & \textbf{0.663}
& \textbf{1.951} \\

UniPhysGen (w/ Image Modality)
& \textbf{78.22} & \textbf{0.435} & \textbf{0.062}
& 0.260 & 0.813 & 0.685 & 0.657
& 1.981 \\
\bottomrule
\end{tabular}
}

\vspace{-4pt}

\end{table*}

\begin{table*}[htbp]
\centering
\caption{Scaling analysis of UniPhysGen with different model sizes.}
\label{tab:scale_res}
\resizebox{\textwidth}{!}{
\begin{tabular}{@{}lcccc|cc|ccc|cccc|c@{}}
\toprule

\multirow{3}{*}{\makecell[lt]{\\ \textbf{Method}}}

& \multicolumn{6}{c|}{\textbf{Articulation Semantics}}
& \multicolumn{8}{c}{\textbf{Physical Properties}} \\

\cmidrule(lr){2-7}
\cmidrule(lr){8-15}

& \multicolumn{4}{c|}{\textbf{Kinematic Parameters}}
& \multicolumn{2}{c|}{\textbf{Articulation Structure}}

& \multicolumn{3}{c|}{\textbf{Material Properties}}
& \multicolumn{4}{c|}{\textbf{Scale \& Mass Properties}}
& \multicolumn{1}{c}{\textbf{Affordance}} \\

\cmidrule(lr){2-5}
\cmidrule(lr){6-7}
\cmidrule(lr){8-10}
\cmidrule(lr){11-14}
\cmidrule(lr){15-15}

& Joint Type
& Axis
& Pivot
& Limit

& Struct.
& Struct.

& Mat.
& $\rho$
& $\mu$

& Dim.
& Dim.
& Mass
& Mass

& Aff. \\

& Acc $\uparrow$
& Ang.\ Err ($^\circ$) $\downarrow$
& Dist.\ Err $\downarrow$
& mIoU $\uparrow$

& mIoU $\uparrow$
& F1 $\uparrow$

& Acc $\uparrow$
& ALDE $\downarrow$
& MAE $\downarrow$

& ALDE $\downarrow$
& MnRE $\uparrow$
& ALDE $\downarrow$
& MnRE $\uparrow$

& MAE $\downarrow$ \\

\midrule

UniPhysGen-0.6B
& 89.19 & \textbf{8.66} & 0.110 & \textbf{85.36}
& {78.71} & {80.30}

& {74.89} & {0.475} & {0.064}
& {0.283} & {0.801} & {0.718} & {0.643}
& {2.108} \\

UniPhysGen-1.7B
& \textbf{89.96} & 9.77 & \textbf{0.099} & 84.95
& \textbf{80.46} & \textbf{83.21}

& \textbf{77.92} & \textbf{0.438} & \textbf{0.062}
& \textbf{0.251} & \textbf{0.819} & \textbf{0.664} & \textbf{0.663}
& \textbf{1.951} \\

\bottomrule
\end{tabular}
}
\footnotesize
\end{table*}


\section{Conclusion}
In this paper, we introduce UniPhys, a scalable pipeline for unified physical grounding of heterogeneous 3D assets, enabling the transformation of raw meshes into simulation-ready assets.
To support this direction, we construct UniPhys-40K and a manually verified benchmark UniPhys-Bench. Based on these resources, we further propose UniPhysGen, a unified physical grounding model that jointly reasons over articulation semantics and intrinsic physical properties across diverse 3D assets.
Extensive experiments demonstrate strong performance and robustness on both articulation grounding and physical property estimation, and enable realistic physical interaction in simulation environments. We hope UniPhys can serve as a foundation for physically grounded 3D understanding and embodied interaction.

\bibliography{main}

@inproceedings{xu2025gaussianproperty,
  title={Gaussianproperty: Integrating physical properties to 3d gaussians with lmms},
  author={Xu, Xinli and Ge, Wenhang and Qiu, Dicong and Chen, ZhiFei and Yan, Dongyu and Liu, Zhuoyun and Zhao, Haoyu and Zhao, Hanfeng and Zhang, Shunsi and Liang, Junwei and others},
  booktitle={Proceedings of the IEEE/CVF International Conference on Computer Vision},
  pages={7231--7240},
  year={2025}
}

@inproceedings{zhai2024physical,
  title={Physical property understanding from language-embedded feature fields},
  author={Zhai, Albert J and Shen, Yuan and Chen, Emily Y and Wang, Gloria X and Wang, Xinlei and Wang, Sheng and Guan, Kaiyu and Wang, Shenlong},
  booktitle={Proceedings of the IEEE/CVF Conference on Computer Vision and Pattern Recognition},
  pages={28296--28305},
  year={2024}
}

@inproceedings{zhao2025real2code,
  title={Real2code: Reconstruct articulated objects via code generation},
  author={Zhao, Mandi and Weng, Yijia and Bauer, Dominik and Song, Shuran},
  booktitle={International Conference on Learning Representations},
  volume={2025},
  pages={668--686},
  year={2025}
}

@article{qiu2025articulate,
  title={Articulate anymesh: Open-vocabulary 3d articulated objects modeling},
  author={Qiu, Xiaowen and Yang, Jincheng and Wang, Yian and Chen, Zhehuan and Wang, Yufei and Wang, Tsun-Hsuan and Xian, Zhou and Gan, Chuang},
  journal={arXiv preprint arXiv:2502.02590},
  year={2025}
}

@inproceedings{li2026particulate,
  title={Particulate: Feed-forward 3d object articulation},
  author={Li, Ruining and Yao, Yuxin and Zheng, Chuanxia and Rupprecht, Christian and Lasenby, Joan and Wu, Shangzhe and Vedaldi, Andrea},
  booktitle={Proceedings of the IEEE/CVF Conference on Computer Vision and Pattern Recognition},
  pages={27708--27718},
  year={2026}
}

@inproceedings{le2025articulate,
  title={Articulate-anything: Automatic modeling of articulated objects via a vision-language foundation model},
  author={Le, Long and Xie, Jason and Liang, William and Wang, Hung-Ju and Yang, Yue and Ma, Yecheng Jason and Vedder, Kyle and Krishna, Arjun and Jayaraman, Dinesh and Eaton, Eric},
  booktitle={International Conference on Learning Representations},
  volume={2025},
  pages={17578--17602},
  year={2025}
}

@article{chen2024urdformer,
  title={Urdformer: A pipeline for constructing articulated simulation environments from real-world images},
  author={Chen, Zoey and Walsman, Aaron and Memmel, Marius and Mo, Kaichun and Fang, Alex and Vemuri, Karthikeya and Wu, Alan and Fox, Dieter and Gupta, Abhishek},
  journal={arXiv preprint arXiv:2405.11656},
  year={2024}
}

@inproceedings{xiang2025structured,
  title={Structured 3d latents for scalable and versatile 3d generation},
  author={Xiang, Jianfeng and Lv, Zelong and Xu, Sicheng and Deng, Yu and Wang, Ruicheng and Zhang, Bowen and Chen, Dong and Tong, Xin and Yang, Jiaolong},
  booktitle={Proceedings of the IEEE/CVF conference on computer vision and pattern recognition},
  pages={21469--21480},
  year={2025}
}

@article{xiang2025native,
  title={Native and compact structured latents for 3d generation},
  author={Xiang, Jianfeng and Chen, Xiaoxue and Xu, Sicheng and Wang, Ruicheng and Lv, Zelong and Deng, Yu and Zhu, Hongyuan and Dong, Yue and Zhao, Hao and Yuan, Nicholas Jing and others},
  journal={arXiv preprint arXiv:2512.14692},
  year={2025}
}

@article{zhao2025hunyuan3d,
  title={Hunyuan3d 2.0: Scaling diffusion models for high resolution textured 3d assets generation},
  author={Zhao, Zibo and Lai, Zeqiang and Lin, Qingxiang and Zhao, Yunfei and Liu, Haolin and Yang, Shuhui and Feng, Yifei and Yang, Mingxin and Zhang, Sheng and Yang, Xianghui and others},
  journal={arXiv preprint arXiv:2501.12202},
  year={2025}
}

@article{lai2025hunyuan3d,
  title={Hunyuan3d 2.5: Towards high-fidelity 3d assets generation with ultimate details},
  author={Lai, Zeqiang and Zhao, Yunfei and Liu, Haolin and Zhao, Zibo and Lin, Qingxiang and Shi, Huiwen and Yang, Xianghui and Yang, Mingxin and Yang, Shuhui and Feng, Yifei and others},
  journal={arXiv preprint arXiv:2506.16504},
  year={2025}
}

@article{deitke2023objaverse_xl,
  title={Objaverse-xl: A universe of 10m+ 3d objects},
  author={Deitke, Matt and Liu, Ruoshi and Wallingford, Matthew and Ngo, Huong and Michel, Oscar and Kusupati, Aditya and Fan, Alan and Laforte, Christian and Voleti, Vikram and Gadre, Samir Yitzhak and others},
  journal={Advances in Neural Information Processing Systems},
  volume={36},
  pages={35799--35813},
  year={2023}
}

@inproceedings{deitke2023objaverse,
  title={Objaverse: A universe of annotated 3d objects},
  author={Deitke, Matt and Schwenk, Dustin and Salvador, Jordi and Weihs, Luca and Michel, Oscar and VanderBilt, Eli and Schmidt, Ludwig and Ehsani, Kiana and Kembhavi, Aniruddha and Farhadi, Ali},
  booktitle={Proceedings of the IEEE/CVF conference on computer vision and pattern recognition},
  pages={13142--13153},
  year={2023}
}

@inproceedings{collins2022abo,
  title={Abo: Dataset and benchmarks for real-world 3d object understanding},
  author={Collins, Jasmine and Goel, Shubham and Deng, Kenan and Luthra, Achleshwar and Xu, Leon and Gundogdu, Erhan and Zhang, Xi and Vicente, Tomas F Yago and Dideriksen, Thomas and Arora, Himanshu and others},
  booktitle={Proceedings of the IEEE/CVF conference on computer vision and pattern recognition},
  pages={21126--21136},
  year={2022}
}

@article{fu20213d,
  title={3d-future: 3d furniture shape with texture},
  author={Fu, Huan and Jia, Rongfei and Gao, Lin and Gong, Mingming and Zhao, Binqiang and Maybank, Steve and Tao, Dacheng},
  journal={International Journal of Computer Vision},
  volume={129},
  number={12},
  pages={3313--3337},
  year={2021},
  publisher={Springer}
}

@inproceedings{khanna2024habitat,
  title={Habitat synthetic scenes dataset (hssd-200): An analysis of 3d scene scale and realism tradeoffs for objectgoal navigation},
  author={Khanna, Mukul and Mao, Yongsen and Jiang, Hanxiao and Haresh, Sanjay and Shacklett, Brennan and Batra, Dhruv and Clegg, Alexander and Undersander, Eric and Chang, Angel X and Savva, Manolis},
  booktitle={Proceedings of the IEEE/CVF Conference on Computer Vision and Pattern Recognition},
  pages={16384--16393},
  year={2024}
}

@article{chang2015shapenet,
  title={Shapenet: An information-rich 3d model repository},
  author={Chang, Angel X and Funkhouser, Thomas and Guibas, Leonidas and Hanrahan, Pat and Huang, Qixing and Li, Zimo and Savarese, Silvio and Savva, Manolis and Song, Shuran and Su, Hao and others},
  journal={arXiv preprint arXiv:1512.03012},
  year={2015}
}

@inproceedings{mo2019partnet,
  title={Partnet: A large-scale benchmark for fine-grained and hierarchical part-level 3d object understanding},
  author={Mo, Kaichun and Zhu, Shilin and Chang, Angel X and Yi, Li and Tripathi, Subarna and Guibas, Leonidas J and Su, Hao},
  booktitle={Proceedings of the IEEE/CVF conference on computer vision and pattern recognition},
  pages={909--918},
  year={2019}
}

@inproceedings{xiang2020sapien,
  title={Sapien: A simulated part-based interactive environment},
  author={Xiang, Fanbo and Qin, Yuzhe and Mo, Kaichun and Xia, Yikuan and Zhu, Hao and Liu, Fangchen and Liu, Minghua and Jiang, Hanxiao and Yuan, Yifu and Wang, He and others},
  booktitle={Proceedings of the IEEE/CVF conference on computer vision and pattern recognition},
  pages={11097--11107},
  year={2020}
}

@article{yang2026physforge,
  title={PhysForge: Generating Physics-Grounded 3D Assets for Interactive Virtual World},
  author={Yang, Yunhan and Wang, Chunshi and Ye, Junliang and Li, Yang and Chen, Zanxin and Huang, Zehuan and Mu, Yao and Chen, Zhuo and Guo, Chunchao and Liu, Xihui},
  journal={arXiv preprint arXiv:2605.05163},
  year={2026}
}

@inproceedings{NEURIPS2025_86beeac1,
  author = {Cao, Ziang and Chen, Zhaoxi and Pan, Liang and Liu, Ziwei},
  title = {PhysX-3D: Physical-Grounded 3D Asset Generation},
  booktitle = {Advances in Neural Information Processing Systems},
  pages = {93771--93784},
  volume = {38},
  year = {2025}
}

@article{cao2025physx,
  title={PhysX-Anything: Simulation-Ready Physical 3D Assets from Single Image},
  author={Cao, Ziang and Hong, Fangzhou and Chen, Zhaoxi and Pan, Liang and Liu, Ziwei},
  journal={arXiv preprint arXiv:2511.13648},
  year={2025}
}

@inproceedings{iliash2026s2o,
  title={S2o: Static to openable enhancement for articulated 3d objects},
  author={Iliash, Denys and Jiang, Hanxiao and Zhang, Yiming and Savva, Manolis and Chang, Angel X},
  booktitle={Proceedings of the IEEE/CVF Winter Conference on Applications of Computer Vision},
  pages={6785--6795},
  year={2026}
}

@inproceedings{geng2023gapartnet,
  title={Gapartnet: Cross-category domain-generalizable object perception and manipulation via generalizable and actionable parts},
  author={Geng, Haoran and Xu, Helin and Zhao, Chengyang and Xu, Chao and Yi, Li and Huang, Siyuan and Wang, He},
  booktitle={Proceedings of the IEEE/CVF conference on computer vision and pattern recognition},
  pages={7081--7091},
  year={2023}
}

@inproceedings{liu2025partfield,
  title={Partfield: Learning 3d feature fields for part segmentation and beyond},
  author={Liu, Minghua and Uy, Mikaela Angelina and Xiang, Donglai and Su, Hao and Fidler, Sanja and Sharp, Nicholas and Gao, Jun},
  booktitle={Proceedings of the IEEE/CVF International Conference on Computer Vision},
  pages={9704--9715},
  year={2025}
}

@article{ma2025p3,
  title={P3-sam: Native 3d part segmentation},
  author={Ma, Changfeng and Li, Yang and Yan, Xinhao and Xu, Jiachen and Yang, Yunhan and Wang, Chunshi and Zhao, Zibo and Guo, Yanwen and Chen, Zhuo and Guo, Chunchao},
  journal={arXiv preprint arXiv:2509.06784},
  year={2025}
}

@inproceedings{ravi2025sam,
  title={Sam 2: Segment anything in images and videos},
  author={Ravi, Nikhila and Gabeur, Valentin and Hu, Yuan-Ting and Hu, Ronghang and Ryali, Chaitanya and Ma, Tengyu and Khedr, Haitham and R{\"a}dle, Roman and Rolland, Chloe and Gustafson, Laura and others},
  booktitle={International Conference on Learning Representations},
  volume={2025},
  pages={28085--28128},
  year={2025}
}

@inproceedings{apriori,
  author       = {Rakesh Agrawal and
                  Ramakrishnan Srikant},
  editor       = {Jorge B. Bocca and
                  Matthias Jarke and
                  Carlo Zaniolo},
  title        = {Fast Algorithms for Mining Association Rules in Large Databases},
  booktitle    = {VLDB'94, Proceedings of 20th International Conference on Very Large
                  Data Bases, September 12-15, 1994, Santiago de Chile, Chile},
  pages        = {487--499},
  publisher    = {Morgan Kaufmann},
  year         = {1994},
  bibsource    = {dblp computer science bibliography, https://dblp.org}
}

@article{kuhn1955hungarian,
  title={The Hungarian method for the assignment problem},
  author={Kuhn, Harold W},
  journal={Naval research logistics quarterly},
  volume={2},
  number={1-2},
  pages={83--97},
  year={1955},
  publisher={Wiley Online Library}
}

@article{yan2025x,
  title={X-part: high fidelity and structure coherent shape decomposition},
  author={Yan, Xinhao and Xu, Jiachen and Li, Yang and Ma, Changfeng and Yang, Yunhan and Wang, Chunshi and Zhao, Zibo and Lai, Zeqiang and Zhao, Yunfei and Chen, Zhuo and others},
  journal={arXiv preprint arXiv:2509.08643},
  year={2025}
}

@article{yang2025qwen3,
  title={Qwen3 technical report},
  author={Yang, An and Li, Anfeng and Yang, Baosong and Zhang, Beichen and Hui, Binyuan and Zheng, Bo and Yu, Bowen and Gao, Chang and Huang, Chengen and Lv, Chenxu and others},
  journal={arXiv preprint arXiv:2505.09388},
  year={2025}
}

@inproceedings{wu2025sonata,
  title={Sonata: Self-supervised learning of reliable point representations},
  author={Wu, Xiaoyang and DeTone, Daniel and Frost, Duncan and Shen, Tianwei and Xie, Chris and Yang, Nan and Engel, Jakob and Newcombe, Richard and Zhao, Hengshuang and Straub, Julian},
  booktitle={Proceedings of the Computer Vision and Pattern Recognition Conference},
  pages={22193--22204},
  year={2025}
}

@inproceedings{radford2021learning,
  title={Learning transferable visual models from natural language supervision},
  author={Radford, Alec and Kim, Jong Wook and Hallacy, Chris and Ramesh, Aditya and Goh, Gabriel and Agarwal, Sandhini and Sastry, Girish and Askell, Amanda and Mishkin, Pamela and Clark, Jack and others},
  booktitle={International conference on machine learning},
  pages={8748--8763},
  year={2021},
  organization={PmLR}
}

@article{singh2025openai,
  title={Openai gpt-5 system card},
  author={{OpenAI}},
  journal={arXiv preprint arXiv:2601.03267},
  year={2025}
}

@inproceedings{mao2025spatiallm,
  title={Spatiallm: Training large language models for structured indoor modeling},
  author={Mao, Yongsen and Zhong, Junhao and Fang, Chuan and Zheng, Jia and Tang, Rui and Zhu, Hao and Tan, Ping and Zhou, Zihan},
  booktitle={Advances in Neural Information Processing Systems},
  volume={38},
  pages={45165--45195},
  year={2025}
}
\bibliographystyle{iclr2026_conference}

\clearpage
\appendix
\clearpage
\appendix

\begin{center}
{\Large \textbf{UniPhysGen: Unified Physical Grounding for Simulation-Ready 3D Assets}}\\[10pt]
{\Large Supplementary Material}
\end{center}

\vspace{0.8em}

\noindent
This appendix provides additional details on the UniPhys pipeline, UniPhysGen implementation, evaluation metrics, baseline adaptation, supplementary experiments, prompt templates, and algorithms.

\vspace{0.8em}

\noindent{\large\textbf{Contents}}

\vspace{0.3em}

\begin{itemize}[leftmargin=1.5em,itemsep=3pt]

\item \textbf{A. Details of UniPhys Pipeline}
      \dotfill
      \ref{sec:appendix_uniphys}
    \begin{itemize}[leftmargin=1.4em,itemsep=1pt]
        \item Physically Meaningful Structural Decomposition
              \dotfill
              \ref{sec:appendix_uniphys_1}

        \item Intrinsic Physical Property Grounding
              \dotfill
              \ref{sec:appendix_uniphys_2}

        \item Geometry-aware Articulation Grounding
              \dotfill
              \ref{sec:appendix_uniphys_3}

        \item Simulation-driven Consistency Verification
              \dotfill
              \ref{sec:appendix_uniphys_4}

        \item Dataset Construction Details
              \dotfill
              \ref{sec:appendix_uniphys_5}
    \end{itemize}

\item \textbf{B. Details of UniPhysGen}
      \dotfill
      \ref{sec:appendix_uniphysgen}
    \begin{itemize}[leftmargin=1.4em,itemsep=1pt]
        \item Physical Semantic Alignment Pretraining
              \dotfill
              \ref{sec:appendix_uniphysgen_1}

        \item Geometry-Robust Articulation Grounding
              \dotfill
              \ref{sec:appendix_uniphysgen_2}

        \item Articulation Structure Grounding
              \dotfill
              \ref{sec:appendix_uniphysgen_3}

        \item Object-Level Physical Grounding
              \dotfill
              \ref{sec:appendix_uniphysgen_4}
    \end{itemize}

\item \textbf{C. Implementation Details}
      \dotfill
      \ref{sec:appendix_implementation}

\item \textbf{D. Evaluation Metrics}
      \dotfill
      \ref{sec:appendix_metrics}

\item \textbf{E. Baseline Adaptation and Fairness Protocols}
      \dotfill
      \ref{sec:appendix_fairness}

\item \textbf{F. More Results}
      \dotfill
      \ref{sec:appendix_res}
    \begin{itemize}[leftmargin=1.4em,itemsep=1pt]
        \item Results on PartNet-Mobility
              \dotfill
              \ref{sec:appendix_res_pm}

        \item More Qualitative Comparisons
              \dotfill
              \ref{sec:appendix_res_quality}
    \end{itemize}

\item \textbf{G. Prompt Templates}
      \dotfill
      \ref{sec:appendix_prompts}
    \begin{itemize}[leftmargin=1.4em,itemsep=1pt]
        \item Intrinsic Physical Property Grounding Prompt
              \dotfill
              \ref{sec:basic_prompt}

        \item Geometry-aware Articulation Grounding Prompt
              \dotfill
              \ref{sec:articulation_prompt}
    \end{itemize}

\item \textbf{H. Algorithms}
      \dotfill
      \ref{sec:algorithms}

\end{itemize}

\clearpage

\section{Details of UniPhys Pipeline}
\label{sec:appendix_uniphys}
\subsection{Physically Meaningful Structural Decomposition}
\label{sec:appendix_uniphys_1}

Here we provide additional implementation details for the perceptually guided decomposition framework. Given a mesh $\mathcal{M}$, the objective is to recover a physically grounded decomposition $\mathcal{D}^*$ that preserves articulation-sensitive structures, material-aware regions, and functional connectivity for downstream physical grounding.

The decomposition pipeline consists of four stages.

\textbf{(1) Multi-view Perceptual Prior Construction.}
We first render the mesh under multiple viewpoints and establish dense face-level correspondences through face-ID rendering:
\[
\phi : p \rightarrow f_i,
\]
where $p$ denotes a rendered pixel and $f_i \in \mathcal{F}$ denotes the associated mesh face.
Applying SAM~\cite{ravi2025sam} on rendered images produces view-wise segmentations
\[
\{\mathcal{S}_v\}_{v=1}^{V},
\]
which are subsequently lifted into mesh space through $\phi$ to construct perceptual grouping priors $\mathcal{P}$.

\textbf{(2) Cross-view Part Consensus.}
To estimate the number of physically meaningful structures shared across observations, we aggregate lifted segmentations across views through mesh-level overlap analysis.
Specifically, pairwise IoU is computed between lifted mask regions in mesh space, and regions with IoU larger than $0.5$ are merged into the same perceptual group.
The resulting cross-view perceptual consensus produces an initial part cardinality prior $K$.

\textbf{(3) Candidate Decomposition Alignment.}
We adopt PartField~\cite{liu2025partfield} as the geometry-aware decomposition backbone due to its continuous feature-field formulation, which enables flexible granularity control while preserving local geometric consistency across structurally heterogeneous meshes.
Compared with discrete clustering-based decomposition methods, PartField provides substantially more stable controllable abstraction under varying structural complexities.

Based on the estimated prior $K$, candidate decomposition granularities are generated within a local search range:
\[
\{\mathcal{D}_k\}_{k=K-(K-1)}^{K+15}.
\]

For each candidate decomposition $\mathcal{D}_k$, we evaluate its consistency with the perceptual grouping prior $\mathcal{P}$ by performing Hungarian matching~\cite{kuhn1955hungarian} independently for each view:

\[
\mathcal{L}_{\mathrm{match}}(\mathcal{D}_k)
=
\frac{1}{V}
\sum_{v=1}^{V}
\mathrm{Match}
\!\left(
\mathcal{S}_v,
\mathcal{D}_k^{(v)}
\right),
\]
where $\mathcal{S}_v$ denotes the lifted SAM regions in the $v$-th view, $\mathcal{D}_k^{(v)}$ denotes the candidate decomposition components restricted to the mesh faces visible in that view, and $\mathrm{Match}(\cdot)$ computes the Hungarian matching cost under the optimal one-to-one assignment between the two sets of regions.

The candidate decomposition with the lowest cross-view matching cost is selected:
\[
\mathcal{D}_{\mathrm{match}}
=
\arg\min_{\mathcal{D}_k}
\mathcal{L}_{\mathrm{match}}(\mathcal{D}_k).
\]

\textbf{(4) Structural Refinement.}
To further eliminate over-segmentation, we exploit the consistency of merge evidence across multiple views.
For each view, candidate decomposition components that exhibit sufficient overlap with the same lifted SAM region are grouped into a transaction, yielding a collection of multi-view merge observations.
Apriori-based frequent itemset mining~\cite{apriori} is then applied to discover stable co-occurring merge groups shared across views.
To improve robustness, redundant subset patterns are removed and highly overlapping frequent groups are further consolidated.
The resulting merge groups $\mathcal{M}_{\mathrm{freq}}$ are finally applied to refine the selected decomposition:
\[
\mathcal{D}^{*}
=
\mathcal{R}
\left(
\mathcal{D}_{\mathrm{match}},
\mathcal{M}_{\mathrm{freq}}
\right).
\]
The refined decomposition preserves articulation-sensitive structures while reducing view-specific over-segmentation, providing reliable structural support for downstream articulation grounding and intrinsic physical property estimation.



\begin{figure}[htbp]
  \includegraphics[width=\linewidth]{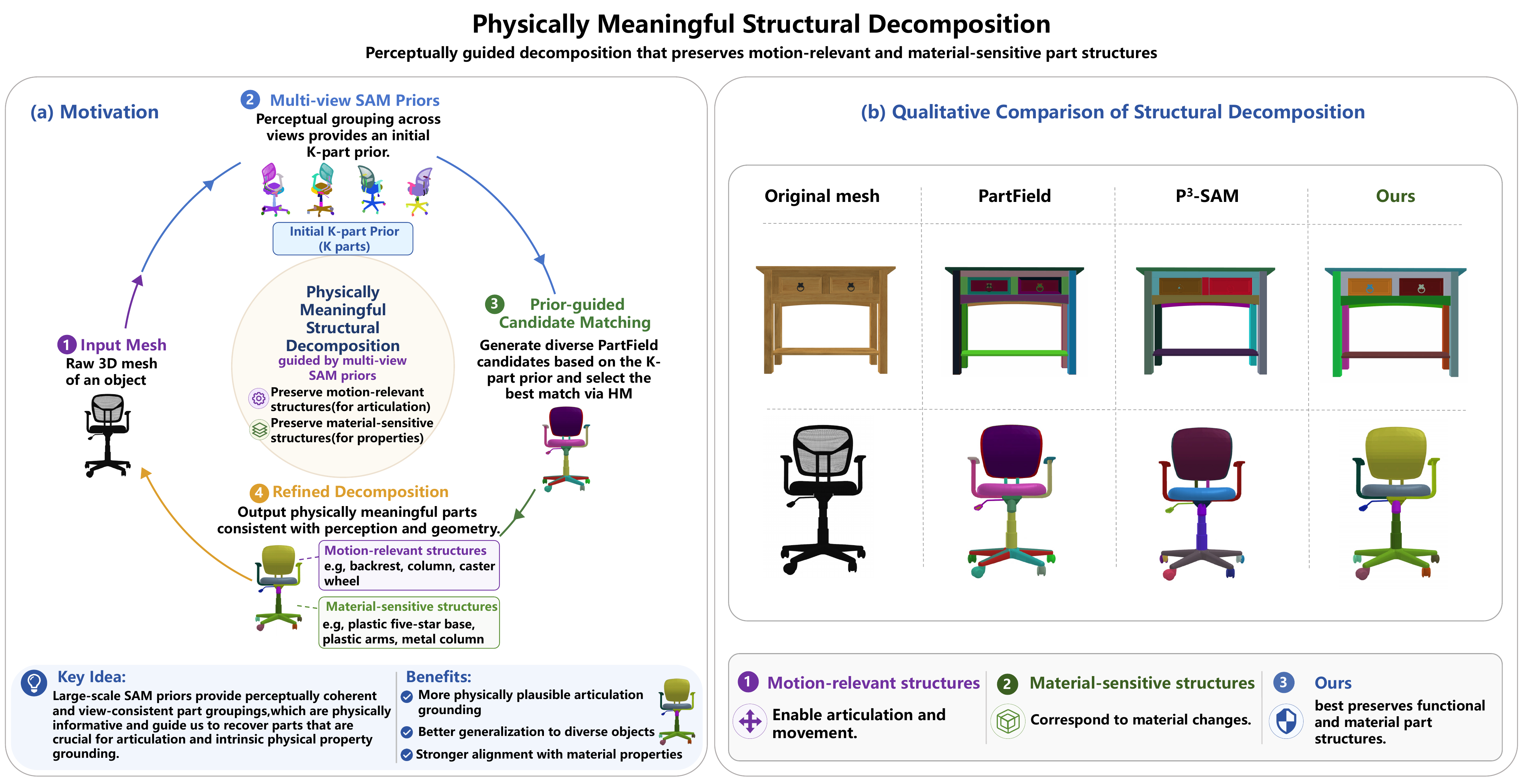}
  \setlength{\abovecaptionskip}{-6pt}
  \caption{Details of Perceptually guided physically meaningful structural decomposition pipeline.
}

  \label{fig:decomp_cmp}
\end{figure}

Algorithm~\ref{alg:decomposition} summarizes the overall perceptually guided decomposition pipeline. 
Fig.~\ref{fig:decomp_cmp} further visualizes the key stages of the decomposition process, including multi-view SAM priors, Hungarian-matched candidate selection, and structural refinement, and provides qualitative examples against geometry-centric decomposition methods such as PartField~\cite{liu2025partfield} and $P^{3}$-SAM~\cite{ma2025p3}. 
These examples illustrate that the proposed decomposition yields physically meaningful part abstractions for downstream articulation grounding and intrinsic physical property estimation.

\subsection{Intrinsic Physical Property Grounding}
\label{sec:appendix_uniphys_2}

This section provides additional implementation details for the part-aware multimodal physical property grounding stage. Given the physically grounded decomposition $\mathcal{D}^*$, our goal is to infer intrinsic physical properties for both individual parts and the complete object while preserving global structural context.

\begin{figure}[htbp]
  \includegraphics[width=\linewidth]{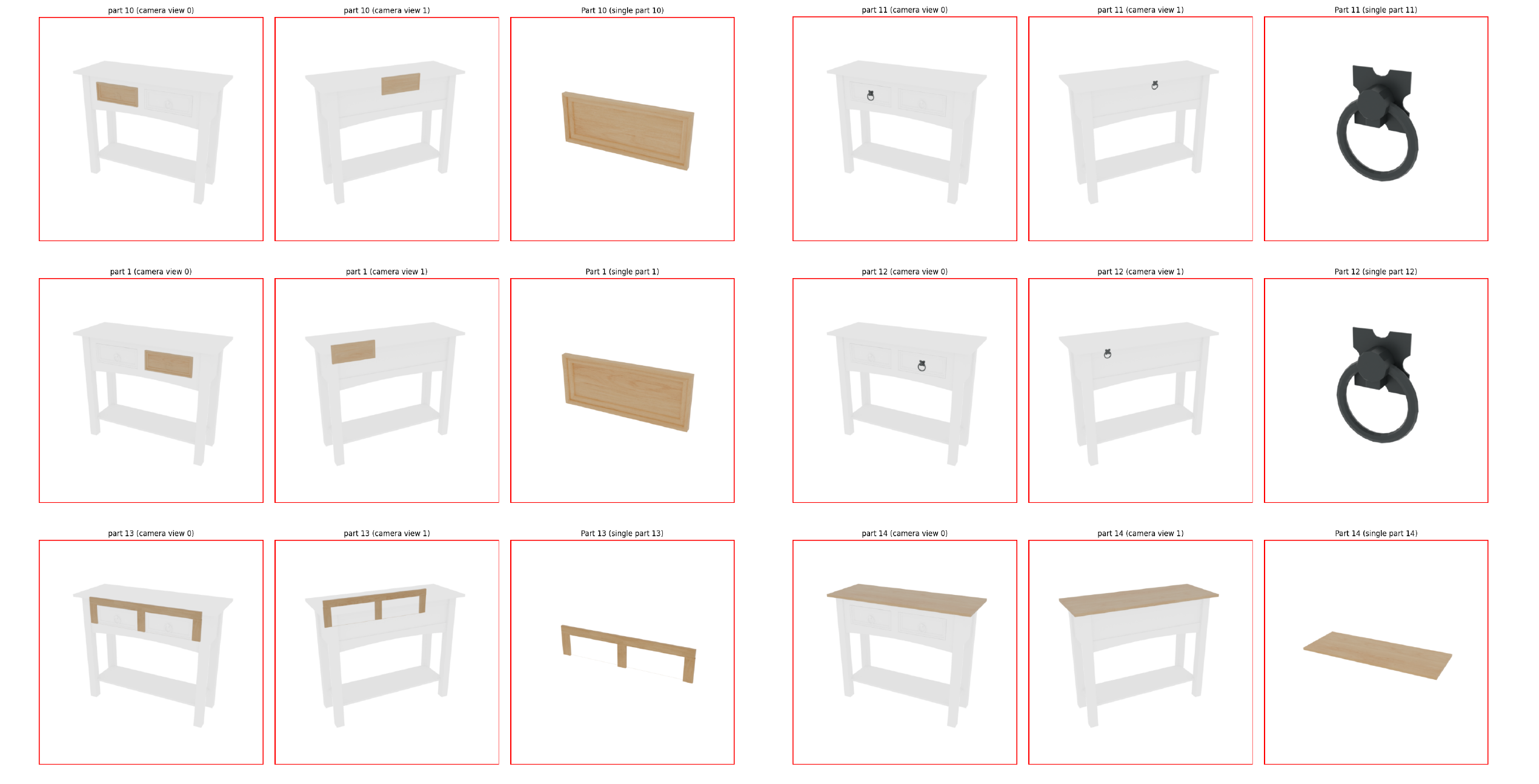}
  \setlength{\abovecaptionskip}{-6pt}
  \caption{Visualization of context-preserving part-centric renderings for intrinsic physical property grounding.
For each decomposed part, the target component is rendered with preserved appearance cues while the surrounding object structure is retained as contextual reference.
Multi-view object observations and localized close-up views are organized into a unified part atlas, enabling the LVLM to jointly reason over material semantics, functional affordance, and global object context.
}

  \label{fig:part_atlas}
\end{figure}

\paragraph{Context-preserving Part-centric Rendering.}
For each decomposed component $d_i \in \mathcal{D}^*$, we construct a set of context-preserving renderings rather than isolated part crops. Specifically, the target part is rendered with its original texture and appearance cues, while the remaining object geometry is retained as a semi-transparent or visually weakened contextual reference. This design allows the LVLM to observe fine-grained material appearance of the target part while maintaining awareness of its global structural role and neighboring components. We further combine multi-view object-level observations with localized close-up renderings to capture both global object semantics and part-level material details. The rendered observations of all decomposed parts are organized into a unified part atlas for LVLM reasoning. 
For clarity, Fig.~\ref{fig:part_atlas} illustrates only a representative subset of part groups from the complete atlas.

\paragraph{LVLM-based Physical Prior Inference.}
We use GPT-5~\cite{singh2025openai} as the multimodal reasoning model to infer structured physical priors from the rendered part atlas. The LVLM is prompted to jointly reason over part identity, material semantics, functional affordance, neighboring structural relations, and object-level physical context. The output is constrained into a structured JSON format containing part-level properties such as material category, density, friction coefficient, elasticity-related attributes, affordance score, and object-level properties such as scale and mass. This structured output format enables automatic parsing and downstream consistency verification.

\paragraph{Prompt Template.}
To improve annotation consistency, we use a fixed prompt template for all assets.
As shown in Listing~\ref{lst:property_prompt}, the prompt specifies the role of the LVLM, the organization of the visual input atlas, the definition of each physical attribute, the required structured output schema, and constraints on physical plausibility, numerical ranges, and part-label consistency.

\subsection{Geometry-aware Articulation Grounding}
\label{sec:appendix_uniphys_3}

\begin{figure}[htbp]
  \includegraphics[width=\linewidth]{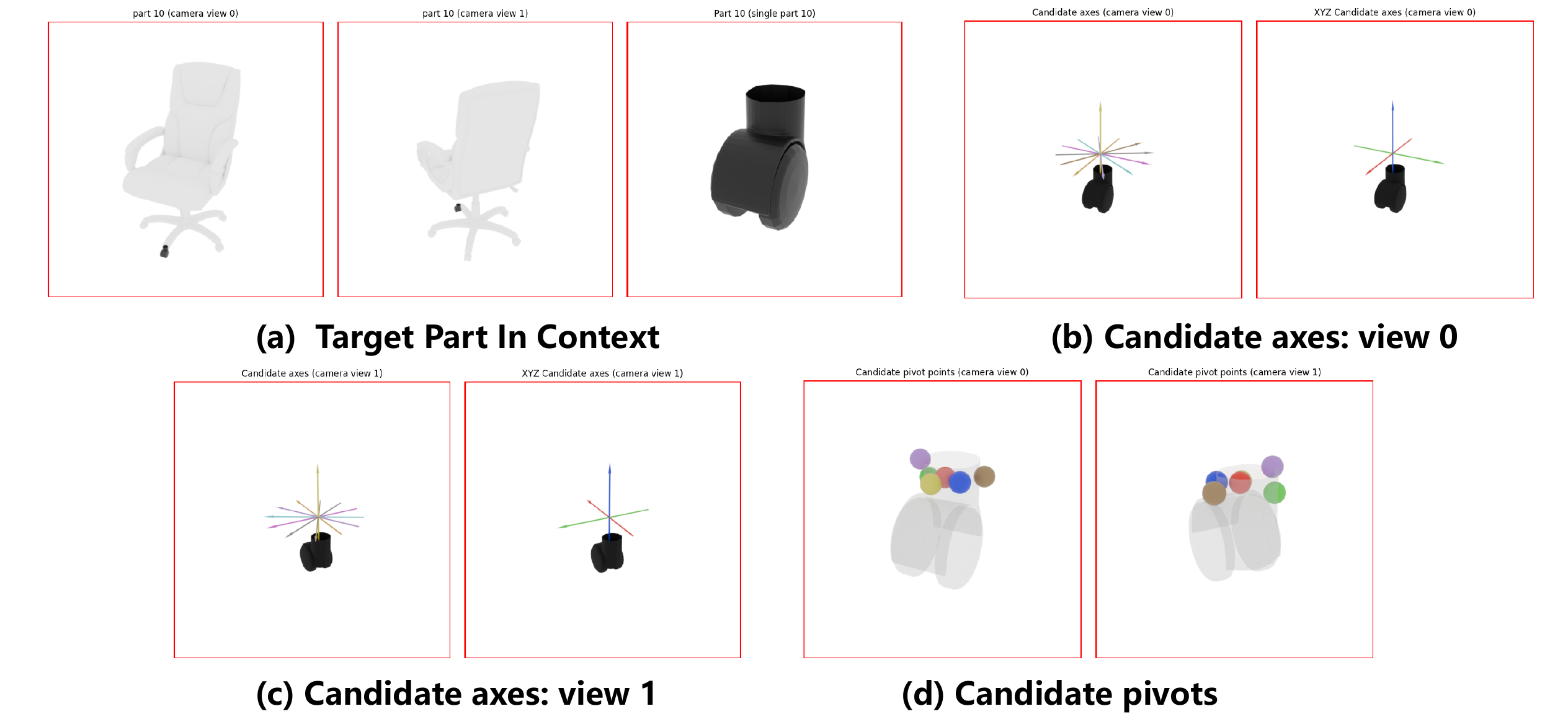}
  \setlength{\abovecaptionskip}{-6pt}
  \caption{Visualization of the LVLM input for geometry-aware articulation grounding.
  For each movable part, UniPhys provides an assembly visualization, two candidate-axis views, and, for revolute joints, pivot-candidate views.
  The candidate axes and pivot points are generated from contact-aware geometric reasoning and rendered as visual-semantic hypotheses, allowing the LVLM to select the physically plausible articulation configuration from a constrained candidate space.
}

  \label{fig:articulation_prompt_vis}
\end{figure}

This section provides additional details for the geometry-aware articulation grounding stage.
Given the motion dependency graph (MDG) constructed from neighboring relations, UniPhys generates geometrically feasible articulation hypotheses for each candidate parent-child motion pair, and then uses the GPT-5~\cite{singh2025openai} to select the most semantically plausible hypothesis from the constrained candidate space.

\paragraph{Geometry-constrained Candidate Generation.}
For each movable component, we first identify the local contact regions between the component and its parent component.
The contact geometry is used to fit a motion-consistent plane, from which candidate articulation axes are derived.
Specifically, we sample six in-plane directions from the fitted motion plane and include the plane normal as an additional candidate.
We further include the three global coordinate axes, i.e., the \(x\)-, \(y\)-, and \(z\)-axes, as valid reference candidates, resulting in ten candidate motion axes in total.
These candidate axes are rendered from two diagonal viewpoints to improve spatial disambiguation.

For revolute joints, UniPhys additionally generates pivot candidates from the contact geometry.
We cluster contact points into five groups and use their cluster centers as candidate pivot points.
We further include the centroid of the entire contact point cloud as an additional candidate, resulting in six pivot candidates.
To make pivot locations visually distinguishable, the target part is rendered semi-transparently while pivot candidates are shown as opaque colored markers.

\paragraph{LVLM-based Semantic Selection.}
The generated articulation hypotheses are rendered into a structured visual input set, as illustrated in Fig.~\ref{fig:articulation_prompt_vis}.
For each movable part, the LVLM (GPT-5~\cite{singh2025openai})receives: (1) an assembly visualization showing the target part in context; (2) two candidate-axis views containing the ten rendered axes; and (3) for revolute joints, two pivot-candidate views.
We additionally provide the motion type and part-level textual descriptions from the intrinsic physical property grounding stage, including basic, functional, and movement descriptions.
This allows the LVLM to select the articulation axis and pivot point by combining geometric feasibility with semantic affordance reasoning.

\paragraph{Prompt Template.}
To improve annotation consistency, we use a fixed prompt template for articulation grounding.
As shown in Listing~\ref{lst:articulation_prompt}, the prompt specifies the LVLM role, the organization of rendered inputs, the candidate-axis and pivot-color conventions, the motion type, the provided part descriptions, and the required structured JSON output schema.

\subsection{Simulation-driven Consistency Verification}
\label{sec:appendix_uniphys_4}

This section provides implementation details for the simulation-driven consistency verification framework. 
Given automatically grounded annotations, the goal is not to recover exact real-world physical quantities, but to determine whether the annotations satisfy internally consistent and simulation-admissible constraints for downstream use.

We instantiate the verification framework with two complementary protocols:
(i) \textit{intrinsic physical property verification}, which checks numerical feasibility, material-aware plausibility, intra-part consistency, structural constraints, and object-level mass consistency; and
(ii) \textit{articulation consistency verification}, which evaluates whether inferred kinematic configurations remain stable during articulated simulation.
Annotations failing these checks are either filtered or refined through a lightweight feedback-based correction mechanism.
The overall verification procedure is summarized in Algorithm~\ref{alg:verification}.


\subsubsection{Intrinsic Physical Property Verification.}
Given a grounded object with decomposed parts $\{p_i\}_{i=1}^{N}$, UniPhys verifies the physical plausibility of part-level properties through four levels.
\paragraph{Level 1: Numerical Feasibility and Material Prior.}
We first check whether each physical attribute lies within a globally admissible numerical range and whether it is compatible with material-specific empirical priors.
For a part assigned to a known material category $m_i$, its density, friction coefficient, Young's modulus, hardness, and Poisson's ratio are required to fall within the corresponding material-dependent plausible intervals.
Parts assigned to the ``other'' material category are excluded from material-specific prior checking.

\paragraph{Level 2: Intra-part Physical Consistency.}
We further check whether physical attributes within a single part are mutually compatible.
For engineering materials where hardness has a meaningful mechanical interpretation, we evaluate the hardness-to-Young's-modulus ratio:
\[
    \frac{H_i}{E_i} \in [10^{-3}, 10^{-1}],
\]
where $H_i$ denotes hardness and $E_i$ denotes Young's modulus.
This rule is used as a empirical plausibility heuristic rather than a strict physical law.
Soft, porous, highly deformable, or anisotropic material categories such as rubber, foam, fabric, leather, and ``other'' are excluded from this check.

\paragraph{Level 3: Structural Consistency Across Parts.}
We impose structure-induced physical constraints derived from the functional role of parts.
For example, highly interactive or graspable components are required to have sufficient surface friction:
\[
    \mu_i \ge 0.2,
\]
where $\mu_i$ denotes the friction coefficient of part $p_i$.
This level verifies whether local physical attributes are consistent with part-level affordance and interaction semantics.

\paragraph{Level 4: Object-level Mass Consistency.}
Finally, object-level mass is used as a global consistency constraint over part-level density and volume annotations.
The mass induced by part-level annotations is estimated as:
\[
    \hat{M}_{\text{parts}} = \sum_{i=1}^{N} \rho_i V_i,
\]
where $\rho_i$ and $V_i$ denote the density and estimated volume of part $p_i$, respectively.
We define the relative mass inconsistency as:
\[
    \epsilon =
    \frac{
    |\hat{M}_{\text{parts}} - M_{\text{obj}}|
    }{
    \max(M_{\text{obj}}, 10^{-6})
    },
\]
where $M_{\text{obj}}$ denotes the object-level mass annotation.
Annotations with $\epsilon > 2$ are considered globally inconsistent and are filtered or sent to feedback-based correction.
This threshold is selected to balance annotation coverage and physical plausibility, rather than to enforce high-precision physical simulation.

\paragraph{Part Volume Estimation.}
Part volumes are estimated through voxelized mesh approximation.
For each part mesh, we voxelize its geometry at a resolution of $128^3$ relative to the axis-aligned bounding box and compute the volume as the number of occupied voxels multiplied by the voxel volume.
A global scaling factor is then applied to align mesh-space volume with real-world object dimensions.
This approximation provides a robust and efficient estimate for large-scale consistency verification.

\subsubsection{Articulation Consistency Verification.}
For articulated annotations, UniPhys validates whether the inferred kinematic configuration remains physically feasible during motion simulation.
Given an articulated part, we simulate its predicted motion in MuJoCo and evaluate three levels of motion consistency.

\paragraph{Level 1: Contact Preservation.}
Let $\mathcal{S}_m$ and $\mathcal{S}_s$ denote the surfaces of the movable component and its adjacent component, respectively.
For a distance threshold $\tau$, the contact region is defined as
\[
\mathcal{P}(\tau)
=
\left\{
\mathbf{x}\in\mathcal{S}_s
\;\middle|\;
\min_{\mathbf{y}\in\mathcal{S}_m}
\|\mathbf{x}-\mathbf{y}\|_2
<
\tau
\right\}.
\]
The initial contact threshold $\tau_0$ is obtained as the minimum threshold that yields a valid contact region.
During simulation, the corresponding threshold $\tau_t$ is recomputed at every simulation step.
The articulated motion is regarded as invalid if
\[
\tau_t>\tau_0+\delta,
\]
where $\delta$ is a tolerance margin.
This criterion indicates that maintaining the original contact requires an unrealistically enlarged contact distance, implying that the articulated parts have effectively lost physical contact.
\paragraph{Level 2: Penetration Stability.}
We detect severe interpenetration during articulated motion by
comparing the thickness of the contact region before and after articulation. 
Given the initial contact point set
$\mathcal{P}_0=\{\mathbf{p}_i\}$,
we estimate its dominant contact plane
\[
\mathbf{n}^{\top}\mathbf{x}+D=0,
\]
using robust plane fitting,
where $\mathbf n$ is the unit plane normal and $D$ is the plane offset.

The contact thickness is defined as the extent of the contact points projected onto the plane normal,
\[
T(\mathcal P)
=
\max_{\mathbf p\in\mathcal P}
(\mathbf n^\top\mathbf p+D)
-
\min_{\mathbf p\in\mathcal P}
(\mathbf n^\top\mathbf p+D).
\]

Let
\[
T_{\mathrm{ori}}
=
T(\mathcal P_0)
\]
denote the initial contact thickness.
During simulation, the contact region is recomputed at every simulation step, producing a contact point set $\mathcal P_t$ and the corresponding thickness ratio
\[
r_t
=
\frac{
T(\mathcal P_t)
}{
T_{\mathrm{ori}}
}.
\]

A motion is regarded as physically implausible if
\[
\operatorname{Percentile}_{95}(r_t)
>
\gamma,
\]
where $\gamma$ is a predefined threshold.
A large thickness ratio indicates excessive expansion of the contact region along the contact normal, suggesting severe geometric interpenetration.
\paragraph{Level 3: Contact Identity Consistency.}
Although local contact locations may vary during articulated motion, the contact region should remain associated with the same physical interface.
For each simulation step, we establish nearest-neighbor correspondences between the initial contact point set $\mathcal{P}_0$ and the current contact point set $\mathcal{P}_t$.
The contact identity overlap is defined as
\[
O_t
=
\frac{
\left|
\left\{
\mathbf{p}\in\mathcal{P}_0
\;\middle|\;
\operatorname{dist}
(\mathbf{p},\mathcal{P}_t)
<
\epsilon
\right\}
\right|
}
{
|\mathcal{P}_0|
},
\]
where $\operatorname{dist}(\mathbf{p},\mathcal{P}_t)$ denotes the Euclidean distance from $\mathbf p$ to its nearest point in $\mathcal P_t$, and $\epsilon$ is a predefined matching tolerance.

The articulation is considered inconsistent if
\[
\operatorname{Percentile}_{10}(O_t)
<
\eta,
\]
where $\eta$ is the minimum acceptable overlap ratio.
A persistently low overlap indicates that the original contact interface has been replaced by unrelated contact regions, suggesting an implausible articulation configuration.

\subsection{Dataset Construction Details}
\label{sec:appendix_uniphys_5}

This section provides additional details on the construction of UniPhys-40K and UniPhys-Bench. UniPhys-40K is built from diverse 3D repositories and further processed by the UniPhys pipeline to obtain unified articulation semantics and intrinsic physical property annotations.

\paragraph{Objaverse-Sketchfab Curation.}
We use the Sketchfab subset of Objaverse-XL~\cite{deitke2023objaverse_xl}, also known as ObjaverseV1~\cite{deitke2023objaverse}, as one of the primary sources of raw 3D assets. Although the assets are collected from a single community-driven platform, this subset contains a large-scale and diverse collection of Creative-Commons-licensed 3D models, covering broad object categories, visual styles, and creation processes, including artist-designed models, procedurally created assets, and real-world 3D scans. To better align the data with household, robotics, and embodied-AI scenarios, we select category-labeled assets related to furniture, containers, appliances, tools, and other interactive indoor objects. We perform category-based filtering over both LVIS-labeled and non-LVIS category annotations, followed by duplicate removal and quality filtering. This filtering process removes empty-category assets, irrelevant outdoor or decorative categories, and geometrically unsuitable meshes, producing a curated subset for downstream physical grounding.

\paragraph{Quality-filtered Repository Assets.}
In addition to Objaverse-Sketchfab, UniPhys-40K also incorporates assets from HSSD~\cite{khanna2024habitat}, 3D-FUTURE~\cite{fu20213d}, ABO~\cite{collins2022abo}, and PartNet~\cite{mo2019partnet}. For Objaverse, HSSD, 3D-FUTURE, and ABO, we adopt quality-filtered assets released by Trellis~\cite{xiang2025structured}, which improves geometric fidelity and removes severely corrupted or incomplete meshes. These assets provide diverse real-world object categories and appearance variations, while the UniPhys pipeline further converts them into physically grounded assets with consistent articulation and intrinsic physical property semantics.

\paragraph{PartNet Texture Recovery.}
PartNet~\cite{mo2019partnet} provides high-quality hierarchical part decompositions based on ShapeNet assets, but the released part meshes often lack the original texture and appearance information required for material-aware physical property grounding. To make PartNet assets suitable for intrinsic physical property annotation, we recover part-level textures from the corresponding ShapeNet models.

Specifically, we first align each PartNet object with its corresponding ShapeNet mesh using the coordinate transformation between the two representations. We then split the ShapeNet mesh into material-aware submeshes according to its OBJ/MTL material references. For each PartNet component, we identify the best-matched ShapeNet submesh based on local geometric overlap in the aligned coordinate system. When UV coordinates and texture maps are available, we transfer the texture to the PartNet component through nearest-surface projection and barycentric UV interpolation; otherwise, we fall back to transferring available material appearance attributes. Finally, the textured components are regrouped according to an appropriate PartNet hierarchy level to avoid overly coarse or excessively fragmented decompositions.

This part-aware texture recovery preserves fine-grained component-level appearance cues, which are important for material, density, friction, affordance, and other intrinsic physical property annotations.

\paragraph{UniPhys-Bench Construction.}
To support reliable evaluation, we construct UniPhys-Bench as a carefully verified benchmark containing 1,927 articulated objects with 5,469 motion-relevant components. The benchmark consists of two complementary subsets. The first subset contains 455 objects sampled from the same data distribution as UniPhys-40K but held out exclusively for evaluation and is manually refined with human verification and refinement of articulation annotations and intrinsic physical property plausibility. The second subset contains 1,472 articulated objects provided by Manycore Tech, where part decompositions are created by professional designers. This designer-created subset introduces additional diversity in part granularity, structural organization, and asset modeling style, making the benchmark more representative of heterogeneous real-world 3D assets. For both subsets, articulation parameters and intrinsic physical properties are initially annotated by the UniPhys pipeline and subsequently inspected and corrected by human annotators. Human verification focuses on motion-relevant components, including joint type, motion axis, pivot, motion range, motion-dependent part groupings, and physical property plausibility. UniPhys-Bench therefore provides a rigorous testbed for evaluating unified physical grounding under diverse object categories, structural complexities, and part decomposition styles.


\section{Details of UniPhysGen}
\label{sec:appendix_uniphysgen}

\subsection{Physical Semantic Alignment Pretraining}
\label{sec:appendix_uniphysgen_1}

This section provides additional details on the physical semantic alignment pretraining stage. 
The goal of this stage is to align 3D geometric tokens with structured physical semantic descriptions, so that the model can jointly reason over local part geometry, global object context, articulation behavior, and intrinsic physical properties.

\paragraph{Input Representation.}
For each training sample, the model receives both the target part point cloud and the complete object point cloud. 
The target part provides fine-grained local geometry and material-sensitive shape cues, while the complete object provides global structural context for functional and articulation-aware reasoning. 
Both point clouds, with each point represented by 3D coordinates, RGB color, and surface normal vectors, are encoded using the Sonata point cloud encoder~\cite{wu2025sonata}, and the resulting geometric tokens are projected into the language model embedding space through a lightweight two-layer MLP projector.

\paragraph{Instruction Template.}
During physical semantic alignment pretraining, UniPhysGen follows an instruction-tuning format. 
Given the target part point cloud and the complete object point cloud, the model is prompted to generate a structured physical semantic description of the target part.
The instruction specifies that the output should include part identity, semantic descriptions, articulation behavior, and intrinsic physical properties in a machine-readable JSON-style format.
A compact instruction template is shown in Listing~\ref{lst:pretrain_prompt}.

\begin{lstlisting}[
caption={Full instruction template for physical semantic alignment pretraining.},
label={lst:pretrain_prompt},
basicstyle=\ttfamily\scriptsize,
frame=single,
breaklines=true,
numbers=left,
xleftmargin=1em
]
[PHYSICS]
Estimate part identity, semantic descriptions, and physical properties.

# Multimodal Inputs
<object_point_cloud>
<part_point_cloud>
<image>

# Notes
- affordance: smaller value means higher affordance.
- motion_type:
  A = contact-only (no relative motion),
  B = translation,
  C = rotation,
  D = rigid (fixed, no motion).
- Units: density in g/cm^3 (if applicable), young (Young's modulus) in GPa,
  hardness in HV, poisson (Poisson's ratio) unitless, friction unitless.

# Output (JSON only)
Return a single JSON object with the following schema (no extra keys):
{
  "part_identity": {
    "part_name": string|null,
    "motion_type": "A"|"B"|"C"|"D"|null,
  },
  "semantic_description": {
    "basic": string|null,
    "functional": string|null,
    "movement": string|null,
    "grasp": string|null
  },
  "physical_properties": {
    "material": string|null,
    "density": number|null,
    "young": number|null,
    "hardness": number|null,
    "poisson": number|null,
    "friction": number|null,
    "graspable": boolean|null,
    "affordance": 1|2|3|4|5|6|7|8|9|10|null
  }
}
\end{lstlisting}

\paragraph{Structured Physical Semantic Supervision.}
Instead of supervising isolated scalar attributes independently, we organize the target output as a structured physical semantic description.
Each annotation contains three complementary levels:
(i) \textit{part identity}, which specifies the part name and motion type;
(ii) \textit{semantic description}, which describes the part's appearance, function, movement behavior, and human interaction mode; and
(iii) \textit{physical properties}, which include material category, density, Young's modulus, hardness, Poisson's ratio, friction coefficient, graspability, and affordance score.
This hierarchical supervision format encourages the model to learn correlations among geometry, function, articulation behavior, and material-dependent physical properties within a shared physical semantic space.
Listing~\ref{lst:semantic_label} shows an example of the structured physical semantic supervision used during pretraining.
The output is represented in a JSON-style format for automatic parsing while preserving open-ended semantic descriptions.

\begin{lstlisting}[
caption={Example structured physical semantic supervision used for physical semantic alignment pretraining.},
label={lst:semantic_label},
basicstyle=\ttfamily\scriptsize,
breaklines=true,
frame=single,
numbers=left,
xleftmargin=1em
]
{
  "part_identity": {
  "part_name": "Door Window Glass",
  "motion_type": "D"
    },
    "semantic_description": {
      "basic": "Circular soda-lime glass window mounted in the door; diameter ~30 cm, thickness ~0.6 cm.",
      "functional": "Provides visibility into the drum while maintaining a barrier to water and detergent.",
      "movement": "Rigidly fixed to the door frame; moves only when the door rotates.",
      "grasp": "Not grasped; occasionally touched during cleaning."
    },
    "physical_properties": {
      "material": "glass/Soda_Lime_Glass",
      "density": 2.5,
      "young": 70.0,
      "hardness": 550.0,
      "poisson": 0.23,
      "friction": 0.5,
      "graspable": false,
      "affordance": 6
    }
}
\end{lstlisting}

By learning to generate such structured descriptions from part-level and object-level geometry, the model establishes a unified physical grounding representation that is subsequently shared by downstream tasks, including kinematic parameter estimation, articulation structure grounding, and object-level physical property estimation.

\subsection{Geometry-Robust Articulation Grounding}
\label{sec:appendix_uniphysgen_2}

While large vision-language models possess strong semantic priors for reasoning about object functionality, articulation grounding remains particularly sensitive to geometry shortcut learning and rotational bias. Unlike intrinsic physical property grounding, articulation parameter estimation requires motion-aware reasoning over geometrically equivariant structures, making it substantially more challenging under heterogeneous and incomplete 3D assets.

\subsubsection{View-Dependent Bias under Canonical Training}

We first observe that commonly adopted canonical augmentation strategies based on upright indoor priors introduce strong view-dependent bias during articulation reasoning. Existing approaches typically assume the input object is aligned with the global upright direction and only apply rotation augmentation around the vertical axis:
\begin{equation}
\mathbf{p}' = \mathbf{R}_{z}(\theta)\mathbf{p},
\label{eq:z_aug_app}
\end{equation}
where $\mathbf{p}\in\mathbb{R}^3$ denotes a point coordinate and $\mathbf{R}_{z}(\theta)$ represents a rotation matrix around the $z$-axis.

Under this setting, articulation axis prediction is effectively constrained to a low-dimensional subspace on the unit circle $S^1$. Although such augmentation performs well under canonical indoor configurations, we observe severe degradation under structure-degenerate cases frequently appearing in 3D asset repositories. For example, incomplete drawer assets containing only front panels are often incorrectly predicted as rotational door-like structures due to shortcut correlations learned from canonical upright priors.

\subsubsection{SO(3)-Based Rotation-Robust Articulation Learning}

To mitigate geometry shortcut learning, we remove canonical orientation assumptions and introduce SO(3)-based rotation augmentation:
\begin{equation}
\mathbf{p}' = \mathbf{R}\mathbf{p},
\quad
\mathbf{R}\in SO(3),
\label{eq:so3_aug_app}
\end{equation}
where $\mathbf{R}$ denotes an arbitrary 3D rotation sampled from the special orthogonal group $SO(3)$.

Compared with Eq.~\ref{eq:z_aug_app}, articulation reasoning is no longer restricted to a planar rotational subspace and instead becomes a spherical reasoning problem over $S^2$, as illustrated in Fig.~\ref{fig:axis_aug}. This significantly improves structure-aware articulation reasoning by forcing the model to rely on motion-relevant geometric semantics rather than view-dependent structural shortcuts.

However, we further observe that naïvely introducing SO(3) augmentation substantially destabilizes articulation axis estimation under arbitrary test-time rotations.



\subsubsection{Spherical Axis Parameterization}

We identify that the instability under SO(3)-based augmentation partly originates from token-level sensitivity in motion direction representation. 
Under arbitrary 3D rotations, articulation axis prediction is lifted from a constrained planar space into a full spherical solution space, substantially increasing the difficulty of open-ended axis generation. 
Directly generating Cartesian directions is particularly sensitive to discrete sign tokens in LLM decoding. 
For example, two normalized directions
\[
\mathbf{a}_1=(0.707,0.707,0), 
\qquad
\mathbf{a}_2=(-0.707,0.707,0)
\]
differ only by a negative sign in the first coordinate, but correspond to substantially different parsed 3D directions. 
Such a small token-level change can lead to a large angular error after parsing, even when the textual supervision loss changes only slightly.

To mitigate this representation instability, we represent normalized articulation axes using spherical coordinates rather than directly generating Cartesian vectors:
\[
\theta = \arccos(z),
\qquad
\phi = \operatorname{atan2}(y,x),
\]
where $\mathbf{a}=(x,y,z)$ denotes a normalized axis direction, $\theta\in[0,\pi]$ is the polar angle, and $\phi\in[0,2\pi)$ is the azimuth angle. 
This angular representation reduces reliance on explicit coordinate sign tokens and provides a smoother output space for LLM-based axis generation under arbitrary object rotations. The corresponding Cartesian direction can be recovered by:
\[
\mathbf{a}=
\begin{bmatrix}
\sin\theta\cos\phi \\
\sin\theta\sin\phi \\
\cos\theta
\end{bmatrix}.
\]

\subsubsection{Global Position Encoding for Pivot Localization}

Accurate pivot estimation further requires consistent spatial reasoning between local part geometry and global object structure. Existing methods typically normalize part-level and object-level coordinates independently before positional encoding:
\begin{equation}
\mathbf{g}_i =
\left\lfloor
\frac{\mathbf{p}_i}{s}
\right\rfloor
-
\min_j
\left(
\left\lfloor
\frac{\mathbf{p}_j}{s}
\right\rfloor
\right),
\label{eq:local_grid}
\end{equation}
where $s$ denotes the voxel grid size.

Although effective for isolated geometry understanding, such local normalization destroys global spatial correspondence between part-level and object-level geometries, leading to unstable pivot localization under heterogeneous decompositions.

To address this issue, we introduce a shared global grid coordinate system jointly applied to both part-level and object-level point clouds:
\begin{equation}
\mathbf{g}_i =
\left\lfloor
\frac{\mathbf{p}_i}{s}
\right\rfloor
-
\mathbf{g}_{\min}^{\text{obj}},
\label{eq:global_grid_app}
\end{equation}
where $\mathbf{g}_{\min}^{\text{obj}}$ is computed globally from the complete object geometry and shared across all part-level representations.


The resulting global grid coordinates are first normalized as
\begin{equation}
\tilde{\mathbf{g}}
=
\frac{\mathbf{g}}{G-1},
\end{equation}
where $G$ denotes the reduced grid resolution after hierarchical voxel pooling.

The normalized coordinates are subsequently encoded using Fourier positional encoding:
\begin{equation}
\gamma(\tilde{\mathbf{g}})
=
\Big[
\tilde{\mathbf{g}},
\sin(\omega_1\tilde{\mathbf{g}}),
\cos(\omega_1\tilde{\mathbf{g}}),
\dots,
\sin(\omega_L\tilde{\mathbf{g}}),
\cos(\omega_L\tilde{\mathbf{g}})
\Big],
\label{eq:fourier}
\end{equation}
where
\[
\omega_l
=
\pi\lambda_l,\qquad
\lambda_l
=
1+\frac{l-1}{L-1}\left(\frac{R}{2}-1\right),
\quad l=1,\ldots,L,
\]
with $L$ denoting the number of frequency bands and $R$ the sampling rate.

Unlike independent local coordinate normalization, the proposed global positional encoding preserves spatial consistency between local part geometry and the complete object structure, substantially improving pivot localization accuracy across heterogeneous 3D assets.

\subsubsection{Instruction and Supervision Format}

UniPhysGen performs kinematic parameter grounding through open-ended text generation rather than a fixed regression head. 
Given the target part point cloud and the complete object point cloud, the model is prompted to infer the complete kinematic parameters—including the joint type (referred to as motion type in our annotation), articulation axis, pivot location, and joint limit (referred to as motion range in our annotation)—in a structured JSON format.
For kinematic parameter grounding, motion types follow our annotation convention: B denotes prismatic translation, while C denotes revolute rotation. 
The axis is represented using spherical angles, while the pivot and motion range are represented as continuous numeric values for subsequent parsing and evaluation.

\begin{lstlisting}[
caption={Full instruction template for kinematic parameter grounding.},
label={lst:motion_prompt},
basicstyle=\ttfamily\scriptsize,
frame=single,
breaklines=true,
numbers=left,
xleftmargin=1em
]
[MOTION]
Estimate the kinematic motion of the target part.

# Multimodal Inputs
<object_point_cloud>
<part_point_cloud>

# Notes
- The input point clouds are AABB-normalized and shifted to [0, 2].
- motion_type uses letters: B = translation (prismatic), C = rotation (revolute).
- axis is represented by spherical coordinates (degrees):
  - theta: polar angle from +Z, integer in [0, 180].
  - phi: azimuth angle around Z, integer in [0, 360).
- pivot is a 3-number point [x, y, z] in the same coordinate frame as the point cloud.
- range is a 2-number array [min, max].
- For C (revolute), range is normalized by $2\pi$.
- For B (prismatic), range is in normalized length units.

# Output (JSON only)
Return a single JSON object with the following schema (no extra keys):
{
  "motion_type": "B"|"C"|null,
  "axis": {"theta": integer, "phi": integer}|null,
  "pivot": [number, number, number]|null,
  "range": [number, number]|null
}
\end{lstlisting}

An example supervision target is shown in Listing~\ref{lst:motion_label}.

\begin{lstlisting}[
caption={Example structured supervision target for kinematic parameter grounding.},
label={lst:motion_label},
basicstyle=\ttfamily\scriptsize,
breaklines=true,
frame=single,
numbers=left,
xleftmargin=1em
]
{
  "motion_type": "C",
  "axis": {
    "theta": 90,
    "phi": 90
  },
  "pivot": [
    0.565,
    0.1032,
    1.1591
  ],
  "range": [
    0.0,
    1.0
  ]
}
\end{lstlisting}

\subsection{Articulation Structure Grounding}
\label{sec:appendix_uniphysgen_3}

This section provides additional details on the articulation structure grounding task. 
Different from kinematic parameter grounding, which predicts joint type, axis, pivot, and limit for a target part, articulation structure grounding predicts the set of parts that move together with the target part during articulation.
This task captures motion-coupled structural dependencies under heterogeneous part decompositions.

\paragraph{Input and Output Format.}
For each target part, the model receives the complete object point cloud, the target part point cloud, optional visual observations, and a list of all part identities with their centroid locations. 
The part list provides an explicit global structural reference, allowing the model to reason over which components are rigidly attached or motion-coupled with the target part.
The output is a subset of input part IDs, where the predicted members must include the target part itself.
This formulation avoids assuming a fixed articulation hierarchy and allows motion-dependent structural grouping under diverse decomposition granularities.

\paragraph{Instruction Template.}
Listing~\ref{lst:group_prompt} shows the instruction template used for articulation structure grounding.

\begin{lstlisting}[
caption={Full instruction template for articulation structure grounding.},
label={lst:group_prompt},
basicstyle=\ttfamily\scriptsize,
breaklines=true,
frame=single,
numbers=left,
xleftmargin=1em
]
[GROUP]
Predict the motion-coupled part group(s): which parts will move together when the target part moves.

# Inputs (placeholders)
- parts(id,position): {{PART_LIST}}

# Multimodal Inputs
<object_point_cloud>
<part_point_cloud>
<image>

# Constraints
- "members" must be selected ONLY from the part ids provided in the input parts list.
- Do NOT generate new ids.
- Each id should appear at most once.
- The output must be a subset of the input part ids.
- "members" must include the target part itself.

# Output (JSON only)
Return a single JSON object with the following schema (no extra keys):
{
  "members": [number]
}
\end{lstlisting}

\paragraph{Supervision Example.}
An example supervision target is shown in Listing~\ref{lst:group_label}. 
In this example, parts 1, 5, and 6 form a motion-coupled group and are expected to move together when the target part is articulated.

\begin{lstlisting}[
caption={Example supervision target for articulation structure grounding.},
label={lst:group_label},
basicstyle=\ttfamily\scriptsize,
breaklines=true,
frame=single,
numbers=left,
xleftmargin=1em
]
{
  "members": [
    1,
    5,
    6
  ]
}
\end{lstlisting}

\subsection{Object-Level Physical Grounding}
\label{sec:appendix_uniphysgen_4}

This section provides additional details on the object-level physical grounding task. 
While previous tasks focus on target-part reasoning, this task estimates holistic physical properties from the complete object geometry and visual appearance.
It encourages UniPhysGen to share the pretrained physical semantic representation across both part-level and object-level grounding.

\paragraph{Input and Output Format.}
Given the complete object point cloud and optional visual observations, the model predicts the object identity, category, global dimensions, and mass.
The dimensions are represented as a three-number array in centimeters, corresponding to the object-level length, width, and height.
The mass is represented in kilograms.
This object-level task complements part-level physical grounding by providing global scale and mass supervision, which is important for simulation initialization and physical consistency checking.

\paragraph{Instruction Template.}
Listing~\ref{lst:object_prompt} shows the instruction template used for object-level physical grounding.

\begin{lstlisting}[
caption={Full instruction template for object-level physical grounding.},
label={lst:object_prompt},
basicstyle=\ttfamily\scriptsize,
breaklines=true,
frame=single,
numbers=left,
xleftmargin=1em
]
[OBJECT_LEVEL]
Estimate object identity and object-level physical properties.

# Multimodal Inputs
<object_point_cloud>
<image>

# Notes
- volume is a 3-number array [L, W, H] in centimeters (cm).
- mass is in kilograms (kg).

# Output (JSON only)
Return a single JSON object with the following schema (no extra keys):
{
  "object_name": string|null,
  "category": string|null,
  "volume": [number, number, number]|null,
  "mass": number|null
}
\end{lstlisting}

\paragraph{Supervision Example.}
An example supervision target is shown in Listing~\ref{lst:object_label}.

\begin{lstlisting}[
caption={Example supervision target for object-level physical grounding.},
label={lst:object_label},
basicstyle=\ttfamily\scriptsize,
breaklines=true,
frame=single,
numbers=left,
xleftmargin=1em
]
{
  "object_name": "Chair",
  "category": "Furniture/SeatingFurniture",
  "volume": [
    60.0,
    60.0,
    95.0
  ],
  "mass": 10.5
}
\end{lstlisting}

\section{Implementation Details}
\label{sec:appendix_implementation}

This section provides additional training and implementation details for UniPhysGen. 
All experiments are conducted on NVIDIA H20 GPUs with approximately 96GB memory per GPU. 
UniPhysGen is built upon Qwen3-1.7B~\cite{yang2025qwen3} and uses Sonata~\cite{wu2025sonata} as the point cloud encoder. 
Following SpatialLM~\cite{mao2025spatiallm}, we train UniPhysGen with full-parameter fine-tuning for all main grounding tasks.
The training configurations for different stages are summarized in Table~\ref{tab:training_config}.

\paragraph{Common Training Configuration.}
Unless otherwise specified, we use a learning rate of $1\times10^{-5}$ with a cosine learning-rate scheduler and a warmup ratio of $0.03$. 
Training is performed with bfloat16 precision and TF32 enabled. 
The per-device batch size is set to 2 with gradient accumulation of 4. 
Thus, the effective batch size is 64 for 8-GPU training and 32 for 4-GPU training.

\paragraph{Training Stages.}
The physical semantic alignment pretraining stage is trained on approximately 370K part-level samples from UniPhys-40K for 6 epochs using 8 H20 GPUs. 
This stage aligns part-level and object-level geometry with structured physical semantic descriptions. 
For downstream task-specific fine-tuning, we use the pretrained checkpoint as initialization. 
Motion parameter grounding is trained for 20 epochs on 4 GPUs, since this task requires joint reasoning over joint type, axis, pivot, and joint limit and empirically converges more slowly. 
Articulation structure grounding and object-level physical grounding are trained for 10 and 6 epochs, respectively, on 4 GPUs.

\paragraph{Image Modality Ablation.}
For the image-modality ablation experiments, we incorporate CLIP ViT-L/14~\cite{radford2021learning} image features through a lightweight projector, as shown in Fig.~\ref{fig:uniphysgen}. 
The CLIP encoder is frozen, and the language model is adapted with LoRA fine-tuning. 
We train the image-enhanced variants for 3 epochs using the same optimization settings unless otherwise specified.

\begin{table}[htbp]
\centering
\small
\caption{Training configurations of UniPhysGen.}
\label{tab:training_config}
\resizebox{\linewidth}{!}{
\begin{tabular}{lccccccc}
\toprule
\textbf{Stage} & \textbf{GPUs} & \textbf{Epochs} & \textbf{LR} & \textbf{Batch / GPU} & \textbf{Grad. Acc.} & \textbf{Global Batch Size} & \textbf{Tuning} \\
\midrule
Physical Semantic Alignment Pretraining 
& 8 $\times$ H20 & 6 & $1\times10^{-5}$ & 2 & 4 & 64 & Full \\
Kinematic Parameter Grounding 
& 4 $\times$ H20 & 20 & $1\times10^{-5}$ & 2 & 4 & 32 & Full \\
Articulation Structure Grounding 
& 4 $\times$ H20 & 10 & $1\times10^{-5}$ & 2 & 4 & 32 & Full \\
Object-Level Physical Grounding 
& 4 $\times$ H20 & 6 & $1\times10^{-5}$ & 2 & 4 & 32 & Full \\
Image-Modality Ablation 
& 4 $\times$ H20 & 3 & $1\times10^{-5}$ & 2 & 4 & 32 & LoRA \\
\bottomrule
\end{tabular}
}
\end{table}

\section{Evaluation Metrics}
\label{sec:appendix_metrics}

This section provides additional details on the evaluation metrics. Since UniPhysGen generates open-ended text outputs that are parsed into structured annotations, we evaluate each task using task-specific structured metrics.

\paragraph{Kinematic Parameter Metrics.}
We adopt a decoupled evaluation protocol for kinematic parameters. 
Following prior articulation evaluation protocols~\cite{le2025articulate}, we report joint type accuracy and axis angular error. 
For pivot localization and limit (motion range) estimation, we use decoupled metrics that separately evaluate pivot-to-axis localization and interval coverage, avoiding the coupling between axis-direction errors, pivot localization errors, and motion-limit errors.

For motion type prediction, we report joint type accuracy. 

For joint type reasoning, let $\hat{t}_i$ and $t_i$ denote the predicted and ground-truth joint type of the $i$-th evaluated component, respectively. 
Joint type accuracy is defined as:
\[
    \mathrm{Acc}_{\mathrm{joint}}
    =
    \frac{1}{N}
    \sum_{i=1}^{N}
    \mathbbm{1}
    \left[
    \hat{t}_i = t_i
    \right],
\]
where $N$ denotes the number of evaluated motion-relevant components and $\mathbbm{1}[\cdot]$ is the indicator function.

For articulation axis estimation, we compute the angular error between the predicted axis $\hat{\mathbf{a}}$ and the ground-truth axis $\mathbf{a}$ as:
\[
    E_{\mathrm{axis}}
    =
    \arccos
    \left(
    \left|
    \frac{
    \hat{\mathbf{a}}^\top \mathbf{a}
    }{
    \|\hat{\mathbf{a}}\|_2 \|\mathbf{a}\|_2
    }
    \right|
    \right)
    \cdot
    \frac{180}{\pi}.
\]
The absolute inner product treats $\mathbf{a}$ and $-\mathbf{a}$ as equivalent, since they represent the same physical articulation axis.



Before computing pivot distances, all point coordinates are transformed into the same AABB-normalized coordinate frame used by the model. 
Let $\mathcal{P}$ denote the complete object point cloud and $\mathcal{N}(\cdot)$ denote the AABB normalization that maps object coordinates into $[-1,1]$. 
We further shift normalized coordinates by the object-level minimum coordinate:
\[
    \widetilde{\mathbf{p}}
    =
    \mathcal{N}(\mathbf{p})
    -
    \min_{\mathbf{q}\in\mathcal{P}}
    \mathcal{N}(\mathbf{q}),
\]
so that evaluated coordinates lie in a shared normalized frame approximately within $[0,2]$.
This normalization removes scale variations across heterogeneous meshes.

For pivot estimation, we evaluate the distance from the predicted pivot point $\widetilde{\hat{\mathbf{c}}}_i$ to the ground-truth articulation axis defined by a normalized ground-truth point $\widetilde{\mathbf{c}}_i$ and direction $\mathbf{a}_i$:
\[
    E_{\mathrm{pivot}}^{(i)}
    =
    \left\|
    (\widetilde{\hat{\mathbf{c}}}_i-\widetilde{\mathbf{c}}_i)
    \times
    \frac{\mathbf{a}_i}{\|\mathbf{a}_i\|_2}
    \right\|_2.
\]
The reported pivot error is:
\[
    \overline{E}_{\mathrm{pivot}}
    =
    \frac{1}{N_{\mathrm{pivot}}}
    \sum_{i=1}^{N_{\mathrm{pivot}}}
    E_{\mathrm{pivot}}^{(i)}.
\]
This metric measures whether the predicted pivot lies on the correct rotation axis, while axis direction is evaluated separately by angular error.

For joint limit estimation, we canonicalize each raw motion range $r=[r_0,r_1]$ into an unsigned interval:
\[
    \widetilde{r}
    =
    \left[
    \min(|r_0|,|r_1|),
    \max(|r_0|,|r_1|)
    \right].
\]
This canonicalization focuses the evaluation on motion-extent coverage rather than sign conventions induced by axis orientation.
We then compute motion range interval IoU between the canonicalized predicted motion range 
$\widetilde{\hat{r}}_i=[\widetilde{\hat{r}}_{i,0},\widetilde{\hat{r}}_{i,1}]$ 
and the canonicalized ground-truth motion range 
$\widetilde{r}_i=[\widetilde{r}_{i,0},\widetilde{r}_{i,1}]$:
\[
    \mathrm{IoU}_{\mathrm{range}}^{(i)}
    =
    \frac{
    \max(0, \min(\widetilde{\hat{r}}_{i,1}, \widetilde{r}_{i,1}) - \max(\widetilde{\hat{r}}_{i,0}, \widetilde{r}_{i,0}))
    }{
    \max(\widetilde{\hat{r}}_{i,1}, \widetilde{r}_{i,1}) - \min(\widetilde{\hat{r}}_{i,0}, \widetilde{r}_{i,0})
    }.
\]
The reported motion range mIoU is obtained by averaging over all evaluated motion components:
\[
    \mathrm{mIoU}_{\mathrm{range}}
    =
    \frac{1}{N_{\mathrm{range}}}
    \sum_{i=1}^{N_{\mathrm{range}}}
    \mathrm{IoU}_{\mathrm{range}}^{(i)}.
\]
Compared with vector-displacement based motion range errors that couple range estimation with axis direction, motion range mIoU directly measures the coverage consistency of motion extents and decouples motion range evaluation from axis estimation.

\paragraph{Articulation Structure Metrics.}
For articulation structure grounding, each sample predicts a set of part IDs that move together with the target part. 
Let $\hat{\mathcal{S}}_i$ and $\mathcal{S}_i$ denote the predicted and ground-truth motion-coupled part sets for sample $i$, respectively. 
We compute the per-sample set IoU as:
\[
    \mathrm{IoU}_{i}
    =
    \frac{
    |\hat{\mathcal{S}}_i \cap \mathcal{S}_i|
    }{
    |\hat{\mathcal{S}}_i \cup \mathcal{S}_i|
    }.
\]
The reported structure mIoU is obtained by averaging $\mathrm{IoU}_{i}$ over all evaluated samples:
\[
    \mathrm{mIoU}_{\mathrm{struct}}
    =
    \frac{1}{N}
    \sum_{i=1}^{N}
    \mathrm{IoU}_{i}.
\]

For structure F1, we report micro-F1 by aggregating true positives, false positives, and false negatives over all samples:
\[
    \mathrm{TP}
    =
    \sum_{i=1}^{N}
    |\hat{\mathcal{S}}_i \cap \mathcal{S}_i|,
    \qquad
    \mathrm{FP}
    =
    \sum_{i=1}^{N}
    |\hat{\mathcal{S}}_i \setminus \mathcal{S}_i|,
    \qquad
    \mathrm{FN}
    =
    \sum_{i=1}^{N}
    |\mathcal{S}_i \setminus \hat{\mathcal{S}}_i|.
\]
The micro-precision, micro-recall, and micro-F1 are computed as:
\[
    P_{\mathrm{micro}}
    =
    \frac{\mathrm{TP}}{\mathrm{TP}+\mathrm{FP}},
    \qquad
    R_{\mathrm{micro}}
    =
    \frac{\mathrm{TP}}{\mathrm{TP}+\mathrm{FN}},
\]
\[
    F1_{\mathrm{micro}}
    =
    \frac{
    2P_{\mathrm{micro}}R_{\mathrm{micro}}
    }{
    P_{\mathrm{micro}}+R_{\mathrm{micro}}
    }.
\]
When both the predicted and ground-truth sets are empty, the corresponding per-sample IoU is defined as 1.



\paragraph{Intrinsic Physical Property Metrics.}
For intrinsic physical property evaluation, we follow the error metrics used in NeRF2Physics~\cite{zhai2024physical} for continuous positive physical quantities, and apply them to material-related and object-level physical properties. 
Given a prediction $\hat{x}$ and ground truth $x$, Absolute Log Difference Error (ALDE) is defined as:
\[
    \mathrm{ALDE}(\hat{x},x)
    =
    |\log \hat{x} - \log x|.
\]
ALDE measures multiplicative deviation in log space and is therefore suitable for physical quantities whose values may vary across different scales.

For ratio-based consistency, we report Min Ratio Error (MnRE) following~\cite{zhai2024physical}:
\[
    \mathrm{MnRE}(\hat{x},x)
    =
    \min
    \left(
    \frac{\hat{x}}{x},
    \frac{x}{\hat{x}}
    \right).
\]
Although originally used for mass estimation, this ratio-based metric is generally applicable to positive scalar physical quantities and reflects scale-invariant multiplicative consistency. 
Its value lies in $(0,1]$, with higher values indicating better agreement.

Specifically, density is evaluated using ALDE. 
For object-level scale and mass estimation, we report both ALDE and MnRE. 
For object scale, both metrics are computed independently along the three spatial dimensions and then averaged across dimensions.
Material classification is evaluated using high-level category accuracy.
Friction estimation and affordance grounding are both evaluated using mean absolute error (MAE), where affordance MAE is computed over the 1--10 affordance scores.

\section{Baseline Adaptation and Fairness Protocols}
\label{sec:appendix_fairness}

To ensure fair comparison across methods with different output formats and supervision assumptions, we adopt the following evaluation protocols.

\paragraph{Real2Code.}
Real2Code~\cite{zhao2025real2code} does not provide official pretrained checkpoints. 
We therefore reproduce the method using the implementation released by Articulate-Anything~\cite{le2025articulate}, following the implementation details described in the original paper~\cite{zhao2025real2code}.

\paragraph{PARTICULATE.}
PARTICULATE~\cite{li2026particulate} jointly predicts articulated part decomposition and articulation parameters. Following its original evaluation protocol, we perform Hungarian matching between predicted and ground-truth articulated parts before computing articulation metrics. Articulation metrics are then computed only over the matched articulated part pairs.

\paragraph{NeRF2Physics.}
NeRF2Physics~\cite{zhai2024physical} predicts voxel-level dense physical properties without explicit part decomposition. To enable part-level evaluation on UniPhys-Bench, we aggregate dense voxel predictions within annotated part regions and compute part-level property estimates before evaluating the corresponding metrics.

\section{More Results}
\label{sec:appendix_res}

\subsection{Results on PartNet-Mobility}
\label{sec:appendix_res_pm}

We further evaluate UniPhysGen on the standard PartNet-Mobility benchmark~\cite{xiang2020sapien} to compare with existing articulation grounding methods under a widely used articulated-object setting. 
For a fair comparison, we fine-tune UniPhysGen on the training split and evaluate on the corresponding test split following the protocol adopted by PARTICULATE~\cite{li2026particulate}.
Since PartNet-Mobility primarily provides articulation annotations, this evaluation only reports articulation-related metrics, including joint type accuracy, axis error, pivot error, and joint limit mIoU.

\begin{table*}[htbp]
\centering
\footnotesize
\setlength{\tabcolsep}{5pt}
\caption{
Quantitative comparison on kinematic parameters estimation on PartNet-Mobility.
}
\label{tab:partnet_mobility}

\begin{tabular}{lcccc}
\toprule

\multirow{2}{*}{\textbf{Method}}
& \multicolumn{4}{c}{\textbf{Motion Parameters}} \\

\cmidrule(lr){2-5}

& Joint
& Axis
& Pivot
& Limit \\

& Acc $\uparrow$
& Ang.\ Err ($^\circ$) $\downarrow$
& Dist.\ Err $\downarrow$
& mIoU $\uparrow$ \\

\midrule

Real2Code
& 59.27 & 69.06 & 0.326 & -- \\

URDFormer
& 70.91 & 10.39 & 0.620 & 83.92 \\

Articulate-Anything$^{*}$
& \textbf{100.00} & \underline{7.97} & 0.471 & 77.20 \\

PARTICULATE
& \underline{100.00} & \textbf{0.52} & \textbf{0.023} & \underline{81.03} \\

\midrule

UniPhysGen (Ours)
& 98.80 & {8.84} & \underline{0.084} & \textbf{91.45} \\

\bottomrule
\end{tabular}
\\
\footnotesize
\textit{Note: $^{*}$ Articulate-Anything relies on strong semantic priors, canonical coordinate conventions, and in-context articulation templates for articulation generation.}
\end{table*}

As shown in Table~\ref{tab:partnet_mobility}, UniPhysGen achieves performance comparable to PARTICULATE on the standard PartNet-Mobility split and obtains better performance on motion limit estimation. 
This result indicates that although UniPhysGen is designed for unified physical grounding over heterogeneous 3D assets rather than being specialized for PartNet-Mobility, it can still achieve competitive articulation grounding performance on a canonical articulated-object benchmark.

We further evaluate orientation robustness by applying non-canonical rotations to the test assets. 
As shown in Table~\ref{tab:partnet_mobility_rot}, PARTICULATE performs strongly under the canonical z-up setting but degrades substantially when the input assets deviate from the canonical orientation. 
In contrast, UniPhysGen maintains more stable performance under rotated inputs. 
This robustness mainly benefits from the SO(3)-based rotation augmentation and spherical-axis parameterization introduced in Sec.~\ref{sec:uniphysgen_2}, which reduce reliance on canonical object orientation and axis-aligned geometric shortcuts.

\begin{table*}[htbp]
\centering
\footnotesize
\setlength{\tabcolsep}{5pt}
\caption{
Orientation robustness under different input up directions.
}
\label{tab:partnet_mobility_rot}

\resizebox{0.82\textwidth}{!}{
\begin{tabular}{@{}l|cccc|cccc@{}}
\toprule

\multirow{2}{*}{\textbf{Up Dir}}
&
\multicolumn{4}{c|}{\textbf{PARTICULATE}}
&
\multicolumn{4}{c}{\textbf{UniPhysGen (Ours)}}
\\

\cmidrule(lr){2-5}
\cmidrule(lr){6-9}

& Joint & Axis & Pivot & Limit
& Joint & Axis & Pivot & Limit
\\

& Acc $\uparrow$ & Ang.\ Err ($^\circ$) $\downarrow$ & Dist.\ Err $\downarrow$ & mIoU $\uparrow$
& Acc $\uparrow$ & Ang.\ Err ($^\circ$) $\downarrow$ & Dist.\ Err $\downarrow$ & mIoU $\uparrow$
\\

\midrule

+Z
& {100.00} & {0.52} & {0.023} & {81.03}
& {98.80} & {8.84} & {0.084} & {91.45}
\\

-Z
& 89.13 & 7.88 & 0.255 & 76.92
& 98.19 & 10.96 & 0.131 & 87.48
\\

+X
& 68.60 & 55.54 & 0.693 & 42.72
& 95.78 & 12.06 & 0.065 & 88.56
\\

-X
& 74.31 & 44.76 & 0.673 & 44.28
& 97.59 & 10.88 & 0.089 & 87.29
\\

+Y
& 73.74 & 50.31 & 0.543 & 72.83
& 97.59 & 10.72 & 0.128 & 88.88
\\

-Y
& 73.74 & 46.67 & 0.639 & 71.43
& 97.61 & 10.96 & 0.131 & 87.48
\\



\bottomrule
\end{tabular}
}

\vspace{0.2cm}
\textit{
Note: Joint Acc denotes motion type classification accuracy.
Axis Err denotes angular error of predicted motion axes.
Pivot Err denotes pivot point distance error.
Limit mIoU evaluates motion range overlap.
}

\end{table*}

\subsection{More Qualitative Comparisons}
\label{sec:appendix_res_quality}

\begin{figure}[t]
  \includegraphics[width=\linewidth]{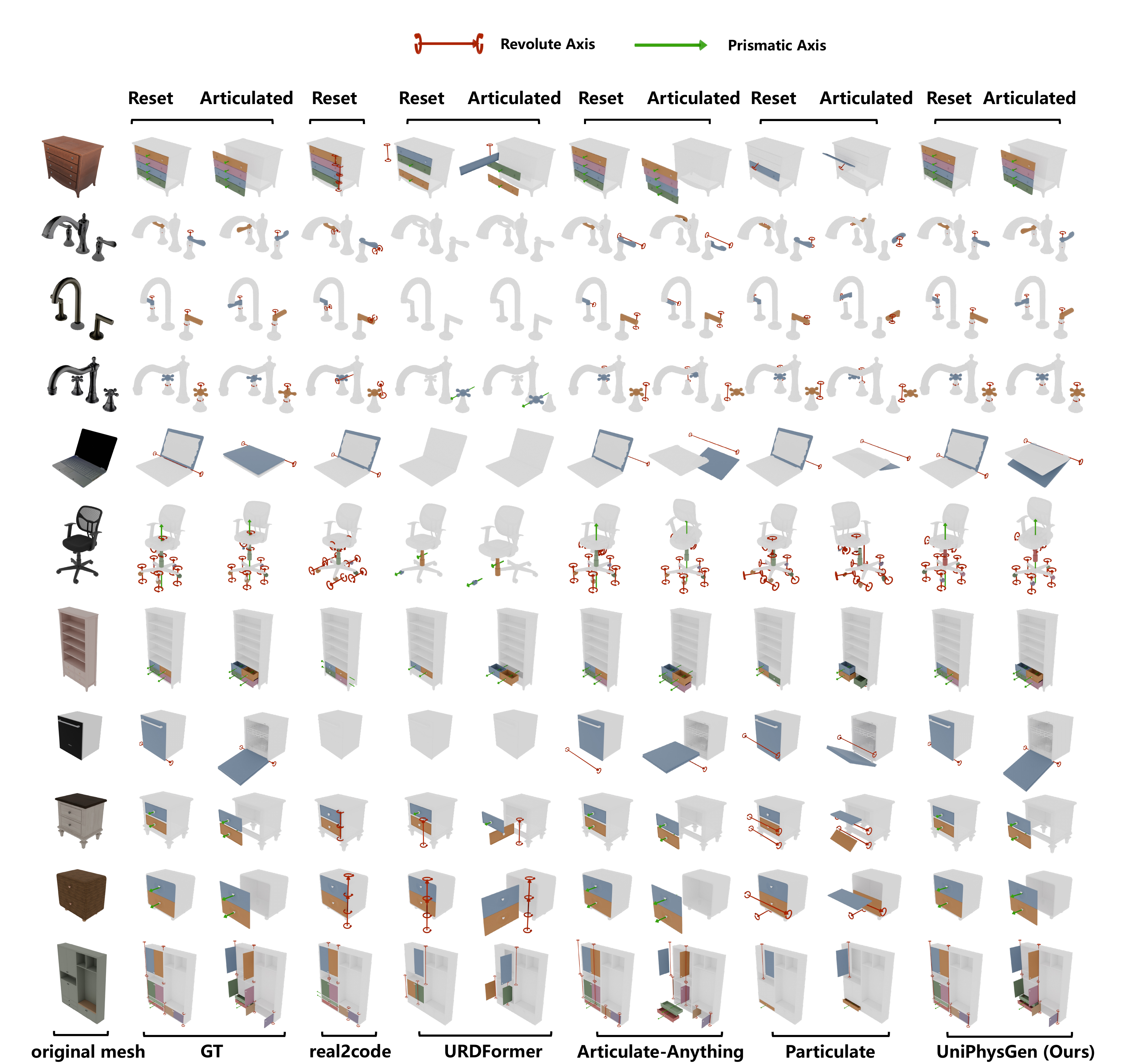}
  \setlength{\abovecaptionskip}{-6pt}
  \caption{Additional qualitative comparisons of articulation grounding on UniPhys-Bench.
“Reset” denotes the initial object state, while “Articulated” denotes the maximally articulated state.
}

  \label{fig:more_arti_quality}
\end{figure}

Fig.~\ref{fig:more_arti_quality} provides additional qualitative comparisons on challenging articulated assets. 
These examples highlight several common failure modes of existing articulation grounding methods and further illustrate the advantages of UniPhysGen under incomplete structures, heterogeneous decompositions, and non-canonical articulated components.

First, incomplete or structure-degenerate assets remain challenging for geometry-centric methods. 
For example, in the drawer-panel cases shown in rows 1, 9, and 10, the asset contains only the drawer front without the complete drawer body. 
Methods that rely heavily on canonical geometric priors often misinterpret such panel-like structures as revolute doors, leading to incorrect rotational motion predictions. 
In contrast, UniPhysGen correctly infers the prismatic drawer motion by jointly reasoning over part-level geometry, functional semantics, and global physical context.

Second, prompt-based methods such as Articulate-Anything~\cite{le2025articulate} can often infer plausible joint types and axis directions when strong semantic cues and canonical coordinate conventions are provided. 
However, as shown in the qualitative results, accurate motion type and axis selection alone do not guarantee physically plausible articulation. 
Errors in pivot localization or overly large motion limits can still lead to unrealistic articulated behavior during simulation. 
This demonstrates the importance of jointly evaluating axis, pivot, and joint limit, rather than relying only on joint type or axis correctness.

Third, purely geometry-grounded methods may fail on components whose motion is determined by functional semantics rather than local shape alone. 
As shown in row 6, caster-like structures require reasoning about the swivel axis through the vertical mounting stem, while wheel-like local geometry may incorrectly suggest rotation around a horizontal axle. 
By aligning geometric representations with structured physical semantics, UniPhysGen better captures such functional motion patterns and produces more physically plausible articulated assets.


\section{Prompt Templates}
\label{sec:appendix_prompts}

\subsection{Intrinsic Physical Property Grounding Prompt.}
\label{sec:basic_prompt}
The full prompt used for LVLM-based physical property grounding is provided below. It defines the composite image input format, fixed part-label ordering, material candidate space, special rules for incomplete geometry, and the structured JSON output schema.

\begin{lstlisting}[
caption={Prompt template for intrinsic physical property grounding.},
label={lst:property_prompt},
basicstyle=\ttfamily\scriptsize,
breaklines=true,
frame=single,
numbers=left,
xleftmargin=1em
]
You have a good understanding of articulated objects and their physical properties.
Your task is to assist the user in analyzing an object from images of its parts, where each image contains three visual regions showing the same part under different views.
You should base your analysis on real-world 3D objects, ensuring that the dimensions, mass, and other physical properties are reasonable and realistic.

=============================
### Input Description
=============================
Each part is represented by ONE composite image that contains THREE visual regions showing the same part under different views:

1. **Left region (camera view 0):** The entire object from one diagonal perspective.
    The target part retains its original color and texture, while all other parts are grayed out, **highlighting the part's position, role, and relationship within the whole object**.

2. **Middle region (camera view 1):** The same object from the back-diagonal perspective corresponding to the Left region.
    Again, the target part remains in original color, and the rest are grayed out, **providing a complementary view of the part's spatial context and function within the object**.

3. **Right region (single part view):** The isolated rendering of the target part alone,
   showing its full geometry and material appearance.

These three regions together describe the same part.
Each composite image corresponds to ONE part, and the order of images strictly defines part labels:

- Image_1 -> Part label = 1
- Image_2 -> Part label = 2
- ...
You must keep this order fixed and consistent throughout the entire analysis and JSON output.
Do NOT reorder or relabel any part based on its function or appearance.

=============================
### 1. Object-level Analysis
=============================
Provide the following properties of the entire object:
(1) name
(2) category
(3) dimension (length*width*height, in cm, e.g., "20*3*2 cm")
(4) mass (in kg, e.g., "0.25 kg")
Ensure the mass and dimensions are consistent with a real-world object of this type.

=============================
### 2. Part-level Analysis
=============================
For each part (e.g., Part_1, Part_2, ...):

Attributes to include:
(1) label (integer)
(2) name
(3) material (choose from candidate list below, formatted as "category/subtype")
(4) density (g/cm$^3$)
(5) Young's Modulus (GPa)
(6) Hardness (HV)
(7) Poisson's Ratio
(8) friction_coefficient (0.2-0.8 typical range; depend on material roughness and contact condition)
(9) priority_rank (1-10, 1 = most important for human interaction; assign based on function)
(10) graspable (True/False)
(11) neighbors: list of neighboring parts with:
     - labels_of_movement_group (e.g., "0-1"): All movement relationships must occur strictly between defined part labels.
     - movement_type:
         A: merely touch, no movement constraints (e.g., two parts only in contact without fixed or movable joint).
         B: relative translation
         C: rotation about an axis
         D: rigid constraint
     - If movement_type = B or C, include:
         (a) parent_label
         (b) child_label
         (c) damping_coefficient (0.01-0.1 typical range; adjust as needed based on motion, material, weight, and connection type)
         (d) movement_range ("0-30 cm" for translation, "0-180 degrees" for rotation)
(12) Basic_description (material, dimensions, category, name)
(13) Functional_description
(14) Movement_description
(15) Grasped_description (how humans interact with this part)

=============================
### 3. Material Candidates
=============================
Choose the high-level material **category** from the list below, and then specify a realistic **subtype**.
The subtypes shown are **examples only**.

High-level categories (example subtypes for reference):
- metal (e.g., Cast_Iron, Aluminum_Alloy, Cobalt_Alloy, Copper_Alloy, Steel, Titanium_Alloy, Stainless_Steel)
- plastic (e.g., ABS_Plastic, PET_Plastic, Nylon_PA66, EVA_Plastic, PP_Plastic, PVC_Plastic, HDPE_Plastic)
- fabric (e.g., Cotton, Linen, Wool, Silk, Polyester, Nylon)
- leather (e.g., Faux_Leather_PU, Full_Grain_Leather, Top_Grain_Leather, Suede_Leather, Patent_Leather, Bonded_Leather)
- foam (e.g., Polyurethane_Foam_Flexible, Polystyrene_Foam_Rigid, PVC_Foam_Rigid, EVA_Foam, Silicone_Foam, Neoprene_Foam)
- rubber (e.g., Natural_Rubber, SBR_Rubber, Neoprene_Rubber, Nitrile_Rubber, Silicone_Rubber)
- glass (e.g., Soda_Lime_Glass, Borosilicate_Glass, Glass_Ceramic)
- stone (e.g., Marble, Limestone, Gypsum, Granite, Slate, Sandstone, Quartzite)
- ceramics (e.g., Brick, Porcelain, Alumina_Ceramic, Zirconia_Ceramic, Silicon_Carbide_Ceramic)
- wood (e.g., Hardwood, Softwood, Bamboo, Plywood, Particle_Board, MDF, Cork)
- concrete (e.g., Plain_Concrete, Reinforced_Concrete, High_Strength_Concrete, Lightweight_Concrete)
- other (e.g., Resin, Gel, Water, Oil, Food, Plant, Bone) Note: For materials not included in the listed categories (metal, plastic, fabric, leather, foam, rubber, glass, stone, ceramics, wood, concrete), including organic, liquid, or soft/gel-like materials.

**Important:**
- Always output in `"category/subtype"` format (e.g., `"wood/Hardwood_Generic"`).
- If none of the example subtypes fit, generate a realistic alternative based on material appearance and function.
- Generate realistic physical material names; do not invent fictional ones.

=============================
### 4. Special Rules
=============================
(1) Treat visually incomplete parts as complete functional objects when estimating their motion type or physical parameters.
    - For example, even if only the front panel of a drawer is visible, it should still be considered a complete drawer with its expected sliding motion.
(2) Do not assume missing geometry implies a broken or immovable object.
(3) Infer motion and physical parameters based on functional semantics, not purely on visible geometry.
(4) For movable parts (movement type B or C), each part may have at most one parent, since a child part is defined to move relative to its parent.
(5) Any neighboring part that moves together with a movable part must either be rigidly connected to it (D-type), or, if it is also a B- or C-type movable part, designated as its child so that its motion is defined relative to the parent.
(6) If a part has any rigid or movable linkage (e.g., movement type D, B, or C) with another part, it must not have an movement type A (contact-only).
(7) When determining neighbor or parent-child relationships, use the Left region as the spatial reference:
    - each part in the Left region must correspond one-to-one to its true physical counterpart.
    - Do not assign neighbors or parent-child links to visually similar but spatially different parts.
    - The Middle region may assist, but note it shows a back diagonal view with left/right and front/back reversed.

=============================
### 5. Color Semantics
=============================
- The **target part** is shown in its **original color and texture**.
- **Non-target parts** are rendered in **gray**, serving only as contextual reference.
- The rightmost sub-image (single part) shows the **isolated geometry** for material inference.
Use all three visual regions together for analysis.

=============================
### 6. Output Requirements
=============================
(1) Output must be a valid JSON enclosed between:
    ===BEGIN_JSON=== and ===END_JSON===
(2) Include both object-level and part-level information.
(3) The output must be strictly machine-readable JSON (no commentary, no markdown).
(4) Each part's material must use "category/subtype" format.
(5) Prioritize physical realism and functional reasoning over aesthetics.
(6) Assign priority_rank based on human interaction importance, not visual prominence.

=============================
### 7. Output Example
=============================
===BEGIN_JSON===
{
  "object_name": "Folding Knife",
  "category": "Tool",
  "dimension": "20*3*2 cm",
  "mass": "0.25 kg",
  "parts": [
    {
      "label": 1,
      "name": "Blade",
      "material": "metal/Steel_Generic",
      "density": "7.8 g/cm^3",
      "Young's Modulus (GPa)": 210,
      "Hardness (HV)": 250,
      "Poisson's Ratio": 0.30,
      "friction_coefficient": 0.35,
      "priority_rank": 2,
      "graspable": false,
      "neighbors": [
        {
          "labels_of_movement_group": "1-3",
          "movement_type": "C",
          "parent_label": 3,
          "child_label": 1,
          "damping_coefficient": 0.04,
          "movement_range": "0-180 degrees"
        }
      ],
      "Basic_description": "This is the blade ...",
      "Functional_description": "It serves as the cutting component...",
      "Movement_description": "It rotates around a pivot axis connected to the handle...",
      "Grasped_description": "Unlikely to be grasped directly due to sharpness."
    },
    {
      "label": 2,
      "name": "Butt",
      "material": "metal/Steel_Generic",
      "density": "7.85 g/cm^3",
      "Young's Modulus (GPa)": 200.0,
      "Hardness (HV)": 240,
      "Poisson's Ratio": 0.30,
      "friction_coefficient": 0.45,
      "priority_rank": 6,
      "graspable": false,
      "neighbors": [
        {
          "labels_of_movement_group": "2-3",
          "movement_type": "D"
        }
      ],
      "Basic_description": "Steel butt at the end of the handle.",
      "Functional_description": "Can be used for striking or balancing the knife.",
      "Movement_description": "Rigidly fixed to the handle.",
      "Grasped_description": "Occasionally touched during reverse grip or impact actions."
    }
    ...
  ]
}
===END_JSON===
\end{lstlisting}

\subsection{Geometry-aware Articulation Grounding Prompt.}
\label{sec:articulation_prompt}
The full prompt used for LVLM-based articulation grounding is provided below. It defines the composite visual input format, candidate-axis and pivot-point encoding, stage-wise reasoning instructions, motion type definitions, and the structured JSON output schema. The prompt ensures the model leverages geometric constraints, contact-aware candidate hypotheses, and part-level context to select physically plausible motion axes and pivot points.

\begin{lstlisting}[
caption={Prompt template for geometry-aware articulation grounding.},
label={lst:articulation_prompt},
basicstyle=\ttfamily\scriptsize,
breaklines=true,
frame=single,
numbers=left,
xleftmargin=1em
]
You are an expert in **3D kinematics and mechanical motion analysis**.
Your task is to assist the user in analyzing movable parts of a 3D object from rendered images, where each image set contains one assembly view and two candidate-axis views.
You should reason based on real-world mechanical joints and plausible physical motion, ensuring that the predicted motion axis ((and, if applicable, a pivot point)) is realistic and consistent with the part's structure and motion type.

=============================
### Input Description
=============================

The given movable part is described by THREE images (or FOUR images for REVOLUTE parts), one motion type and Color-vector mapping of candidate-axis :

1. **Image A (assembly visualization):**

   * A composite image with three panels: the left and middle panels show the entire assembly from two diagonal perspectives, where the **target part retains its original color and texture while all other parts are grayed out**; the right panel shows the **isolated rendering of the target part alone**, revealing its full geometry and material appearance.

2. **Image B (candidate axes view 1):**

   * Contains **ten candidate motion axes** in total:

     * **Seven colored arrows** represent automatically generated axis candidates.
     * **Three X, Y, Z reference arrows** represent the global coordinate directions and **are also valid motion-axis candidates**.
   * The left panel displays the seven candidate axes together.
   * The right panel highlights the X, Y, Z axes for coordinate orientation.
   * The image is Captured from one diagonal perspective.

3. **Image C (candidate axes view 2):**

   * Same content and candidate structure as Image B (seven colored + three XYZ axes), but captured from a **different diagonal perspective**, providing better spatial understanding of axis positions and directions.

4. Image D (Pivot-Point View-only for REVOLUTE parts)

    * Provided only when the motion type is REVOLUTE (rotational).
    * Shows a close-up of the target part only, viewed from two different diagonal perspectives.
    * The part is rendered semi-transparent, while six candidate points are rendered as opaque markers to remain clearly visible.
    * Each point represents a potential pivot point - the possible center of rotation for the part.
    * The six candidate points are color-coded as follows:
        - Red, Green, Blue, Yellow, Purple, Brown.
    * Your task is to select exactly one point that most realistically represents the true rotation center of the part - the location through which the chosen motion axis would pass.
    * If you judge that the best pivot point is the **average of all six points** (i.e., the center among them) rather than any single one, output `"average"` as the chosen point.
    * Otherwise, output the **color** of the selected point (one of: Red, Green, Blue, Yellow, Purple, Brown).

5. **Motion type:**

   * Either REVOLUTE (rotational motion) or PRISMATIC (linear sliding motion), indicating whether the target part primarily rotates or slides relative to its parent assembly.

6. **Color-vector mapping:**

   * A list of candidate-axis colors and their corresponding **3D direction vectors** in global coordinates.
   * This information specifies the spatial orientation of each arrow and must be used when reasoning about the motion axis.
   * These mappings are user-provided for each input case.

7. **Textual Descriptions of the Part:**

   * You are also provided with three short text descriptions that summarize the target part's physical and functional properties:
     * **Basic_description:** A short factual summary of the part's material, dimensions, category or name.
     * **Functional_description:** A concise explanation of what the part does or contributes to the overall assembly (e.g., "connects armrest to seat", "supports rotation of handle").
     * **Movement_description:** A short note describing how the part is expected to move or interact mechanically with its parent or neighbors (e.g., "rotates upward around hinge", "slides outward along guide rail").

   * These textual descriptions are provided to help you better infer the part's realistic motion and mechanical behavior.
   * Use them as **supplementary reasoning cues**, especially when Image A is ambiguous or when multiple candidate axes seem plausible.

=============================
### 1. Task Objective and Reasoning Process
=============================
You must select **exactly one** candidate axis from Images B and C that best explains the target part's motion relative to the assembly, consistent with the given motion type.
When the motion type is REVOLUTE and Image D is provided, also determine the pivot point.

Direction (positive / negative):
In addition to identifying the correct motion axis,
you must also determine the motion direction - whether it is positive or negative relative to the global coordinate system.
Use the following rule:
    * The green, blue, and red reference arrows in the candidate images represent the positive directions of the X, Y, and Z axes respectively.
    * If the part moves toward the positive direction of the chosen axis, label "motion-direction": "positive".
    * If the part moves opposite to that direction (e.g., inward, downward, backward, closing motion), label "motion-direction": "negative".
When reasoning, infer the physically plausible direction of motion (e.g., a drawer opening outward -> positive, a chair backrest reclining backward/downward -> negative).


Follow this Three-stage reasoning process:

(1) Stage one - Part Understanding (from Image A):
    * Analyze Image A and the provided textual descriptions (**Basic_description, Functional_description, Movement_description**) to infer the part's type, function, and potential motion behavior.
    * Determine whether the part most likely rotates (REVOLUTE) or slides (PRISMATIC), and where this motion occurs (e.g., hinge, pivot joint, telescopic rod, wheel, drawer, etc.).
    * Predict one plausible motion direction or axis based on the part's geometry, connections, and role in the assembly.
        * Note: You should consider dependent sub-parts:
            * When identifying the motion axis, always consider attached or co-moving parts (sub-components that move together with the target part).
            * These parts form a functional unit, and their collective motion should guide the correct axis and direction selection.
            * For example, if a chair backrest moves together with a headrest, the chosen motion axis should explain the rotation of both as a combined system.


(2) Stage two - Axis Selection (using Images B & C):
    * Examine Image B and Image C, which display the same set of ten candidate motion axes:
        * Seven colored arrows - automatically generated motion-axis candidates.
        * Three X/Y/Z reference arrows - representing the global coordinate directions and also valid motion-axis candidates.
    * The two images differ only in viewpoint, providing complementary perspectives for better 3D spatial understanding of each axis relative to the part.
    * When the normal vector of the part's main surface is neither parallel nor perpendicular to the ground plane (i.e., it does not align with any global X, Y, or Z axis), give higher priority to the candidate axes shown in the first panel (Seven colored arrows).
    * Compare all ten candidate axes with your Stage 1 prediction.
    * Select exactly one candidate axis whose direction and spatial position best match the real-world motion inferred from Image A and the given motion type.

(3) Stage three - Pivot Point Selection (only for REVOLUTE parts, using Image D)
    * If Image D is provided, identify the most physically plausible pivot point from the six colored candidates.
    * If the real pivot is between them, use the average of all six points as the representative center.
    * This point defines where the chosen motion axis passes through.

Your reasoning should always be physically grounded in real-world mechanical design principles, not just geometric alignment.

=============================
### 2. General Rules
=============================
(1) Treat the highlighted target as a **single functional part** - ignore small subcomponents.
(2) Treat the highlighted object as one unit (part or small subassembly). Do not pick axes based on internal subcomponents' local motions.
(3) Output exactly one axis; when similar, choose the one through the attachment/hinge that best explains the part's overall motion.
(4) Use the assembly context from Image A to understand how the part interacts with neighboring components.
(5) Always output **only one** axis; when multiple are similar, select the most physically meaningful one.
(6) When the motion type is REVOLUTE and Image D is provided, also determine the pivot point.

=============================
### 3. Output Requirements
=============================
Output must include:

(1) **part-description:** What the part is and where it is located in the assembly.
(2) **axis-description:** Direction and approximate location relative to the part.
(3) **axis-color:** The color corresponding to the chosen axis.
(4) **axis-vector:** The 3D vector corresponding to the chosen color (refer to Color-vector mapping).
(5) **motion-direction:** "positive" or "negative", depending on whether the motion proceeds along or opposite to the X/Y/Z axis direction.
(6) **pivot-point:** Only include when Motion Type = REVOLUTE and Image D is provided: the selected pivot as one of "Red", "Green", "Blue", "Yellow", "Purple", "Brown" or "average". For all other cases include "Pivot Point": null.
(7) **justification:** One concise sentence explaining the physical reasoning.

=============================
### 4. Output Format
=============================
The final response should strictly follow this structure:

(1) Output must be a valid JSON enclosed between:
    ===BEGIN_JSON=== and ===END_JSON===
(2) The output must be strictly machine-readable JSON (no commentary, no markdown).


=============================
### 5. Output Example
=============================
--- Example 1
===BEGIN_JSON===
{
    "part-description": "Chair caster wheel assembly mounted at the end of one base leg.",
    "axis-description": "Vertical axis passing through the caster's mounting stem where it swivels relative to the chair base.",
    "axis-color": "Blue",
    "axis-vector": "(0.0, 1.0, 0.0)",
    "motion-direction": "positive",
    "pivot-point": "average",
    "justification": "A swivel caster rotates about a vertical stem at its attachment point, and the blue Y-axis matches this natural pivot direction. Rotate outward along the positive Z direction, hence positive"
}
===END_JSON===

--- Example 2
===BEGIN_JSON===
{
  "part-description": "Office chair backrest assembly attached to the seat base.",
  "axis-description": "Horizontal axis passing through the lower hinge connecting the backrest to the seat frame.",
  "axis-color": "Green",
  "axis-vector": "(1.0, 0.0, 0.0)",
  "motion-direction": "negative",
  "pivot-point": "average",
  "justification": "The backrest reclines backward around the lower hinge, rotating opposite to the positive Y direction, hence negative."
}
===END_JSON===
\end{lstlisting}

\section{Algorithms}
\label{sec:algorithms}
\begin{algorithm}[H]
\caption{Perceptually Guided Physically meaningful structural decomposition}
\label{alg:decomposition}
\small

\KwIn{Mesh $\mathcal{M}$}
\KwOut{Physically grounded decomposition $\mathcal{D}^*$}

Render multi-view face-ID images from $\mathcal{M}$\;

Apply SAM to obtain view-wise segmentations
$\{\mathcal{S}_v\}_{v=1}^{V}$\;

Lift segmentations into mesh space through face correspondence $\phi$\;

Compute cross-view IoU and merge consistent regions\;

Estimate part cardinality prior $K$\;

Generate candidate decompositions
$\{\mathcal{D}_k\}_{k=K-(K-1)}^{K+15}$
using PartField\;

\ForEach{$\mathcal{D}_k$}{
Compute perceptual consistency score
$\mathcal{L}_{match}(\mathcal{D}_k)$
through Hungarian matching\;
}

Select best decomposition:
\[
\mathcal{D}_{match}
=
\arg\min_{\mathcal{D}_k}
\mathcal{L}_{match}(\mathcal{D}_k)
\]
\;

Mine frequent merge patterns
$\mathcal{M}_{freq}$
using Apriori\;

Refine decomposition:
\[
\mathcal{D}^*
=
\mathcal{R}
(
\mathcal{D}_{match},
\mathcal{M}_{freq}
)
\]
\;

\Return{$\mathcal{D}^*$}

\end{algorithm}

\begin{algorithm}[t]
\caption{Simulation-driven Consistency Verification}
\label{alg:verification}
\KwIn{
Grounded asset $\mathcal{G}$ with parts $\{p_i\}_{i=1}^{N}$, part-level physical properties $\{\mathbf{q}_i\}_{i=1}^{N}$, object-level mass $M_{\text{obj}}$, and articulated joints $\mathcal{J}$.
}
\KwOut{
Verified annotation set $\mathcal{G}^{*}$.
}

Initialize verified annotation set $\mathcal{G}^{*} \leftarrow \mathcal{G}$\;

\tcp{Protocol I: Intrinsic Physical Property Verification}
\ForEach{part $p_i \in \mathcal{G}$}{
    Extract material category $m_i$, density $\rho_i$, friction $\mu_i$, Young's modulus $E_i$, hardness $H_i$, and affordance attributes from $\mathbf{q}_i$\;

    \tcp{Level 1: numerical feasibility and material prior}
    \If{$\mathbf{q}_i$ violates global feasible ranges}{
        Mark $p_i$ as physically invalid\;
        Continue\;
    }
    \If{$m_i \neq \texttt{other}$ and $\mathbf{q}_i$ violates material-specific priors}{
        Mark $p_i$ as physically inconsistent\;
        Continue\;
    }

    \tcp{Level 2: intra-part physical consistency}
    \If{$m_i$ is an engineering material and $\frac{H_i}{E_i} \notin [10^{-3},10^{-1}]$}{
        Mark $p_i$ as physically inconsistent\;
    }

    \tcp{Level 3: structural consistency across parts}
    \If{$p_i$ is highly interactive or graspable and $\mu_i < 0.2$}{
        Mark $p_i$ as structurally inconsistent\;
    }
}

\tcp{Level 4: object-level mass consistency}
Estimate part volumes $\{V_i\}_{i=1}^{N}$ through voxelized mesh approximation\;
Compute induced part-level mass $\hat{M}_{\text{parts}} = \sum_i \rho_i V_i$\;
Compute mass inconsistency $\epsilon =
\frac{|\hat{M}_{\text{parts}} - M_{\text{obj}}|}
{\max(M_{\text{obj}},10^{-6})}$\;
\If{$\epsilon > 2$}{
    Send material and density assignments to feedback-based correction or filter the object annotation\;
}

\tcp{Protocol II: Articulation Consistency Verification}
\ForEach{joint $j \in \mathcal{J}$}{
    Initialize articulated simulation in MuJoCo using the predicted joint type, axis, pivot, and motion range\;
    Simulate the articulated motion trajectory $\tau_j$\;

    \tcp{Level 1: contact preservation}
    Recompute the contact region throughout $\tau_j$\;
    \If{the contact region cannot be consistently preserved}{
    Refine or reject the articulation hypothesis of $j$\;
    Continue\;
    }

    \tcp{Level 2: penetration stability}
    Compute contact thickness ratio $\frac{T_{\text{cur}}}{T_{\text{ori}}}$ along $\tau_j$\;
    \If{$\operatorname{Percentile}_{95}\left(\frac{T_{\text{cur}}}{T_{\text{ori}}}\right) > 3$}{
        Refine or reject the articulation hypothesis of $j$\;
        Continue\;
    }

    \tcp{Level 3: contact identity consistency}
    Compute contact overlap score $O_{\text{contact}}$ along $\tau_j$\;
    \If{$\operatorname{Percentile}_{10}(O_{\text{contact}}) < 0.4$}{
        Refine or reject the articulation hypothesis of $j$\;
    }
}

Remove annotations marked invalid after feedback correction\;
\Return{$\mathcal{G}^{*}$}\;
\end{algorithm}

\end{document}